\documentclass[sigconf]{acmart}

\usepackage{booktabs}
\usepackage{todonotes}
\usepackage{multirow}
\usepackage{array}
\usepackage[font=footnotesize,hypcap=true]{caption}
\usepackage[font=footnotesize,hypcap=true]{subcaption}

\makeatletter
\DeclareRobustCommand*\cal{\@fontswitch\relax\mathcal}
\makeatother

\usepackage{dcolumn}
\newcolumntype{d}[1]{D{.}{.}{#1}}
\newcolumntype{e}[1]{D{e}{\textsc{e}\text{-}}{#1}}

\usepackage[inline,shortlabels]{enumitem}
\newlist{inparaenum}{enumerate*}{1}
\setlist*[inparaenum,1]{%
  label=(\roman*),
  ref=\roman*,
}

\let\leftorig\left
\let\rightorig\right
\renewcommand{\left}{\mathopen{}\mathclose\bgroup\leftorig}
\renewcommand{\right}{\aftergroup\egroup\rightorig}

\newcommand{\ccite}[1]{\protect\cite{#1}}

\newcommand{\tsne}{{t-SNE}}     % [not math-mode]
\newcommand{\spmv}{\texttt{SpMV}}
\newcommand{\spmvMkl}{\texttt{MKL\_CSC\_MV}}

\newcommand{\mbs}[1]{\boldsymbol{\mathbf{#1}}}

\DeclareMathOperator{\nnzOp}{nnz}

\newcommand{\nnz}[1]{\nnzOp\left(#1\right)}

\newcommand{\idxIter}{\kappa}

\newcommand{\dataMat}{\mbs{X}}

\newcommand{\permMat}{\mbs{\pi}}
\newcommand{\permTgt}{\permMat_{\text{t}}}
\newcommand{\permSrc}{\permMat_{\text{s}}}
\DeclareMathOperator{\patchDensity}{\beta}
\DeclareMathOperator{\area}{area}
\newcommand{\mat}{\mbs{A}}
\newcommand{\submat}{\mbs{B}}
\newcommand{\idxPatch}{\ell}
\newcommand{\patch}{\submat_{\idxPatch}}

\newcommand{\patchCovering}{\mathcal{P\!C}}
\newcommand{\nonzeros}{\mathcal{I}_{\text{nz}}}

\DeclareMathOperator*{\argmax}{arg \smsp max}

\newcommand{\smsp}{\mspace{2mu}}  % slightly smaller space than \,
\newcommand{\patchDensityKern}{\gamma}
\newcommand{\scale}{\sigma}
\newcommand{\idxA}{\mbs{p}}
\newcommand{\idxB}{\mbs{q}}
\newcommand{\norm}[1]{\left\| #1 \right\|}

\newcommand{\setenum}[1]{\left\{ #1 \right\}}
\newcommand{\setcard}[1]{\left| #1 \right|}
\newcommand{\setcond}[2]{\left\{ #1 \,\middle\vert\, #2 \right\}}

\DeclareMathOperator{\nn}{near-neighbors}
\newcommand{\almost}{{\sim}}

\newcommand\hmm[1]{\ifnum\ifhmode\spacefactor\else2000\fi>1000\uppercase{#1}\else#1\fi}

\setcopyright{rightsretained}
\acmConference{}{}{}

\acmYear{2017}

\copyrightyear{2017}

\title{Rapid Near-Neighbor Interaction of 
  High-dimensional Data 
  via Hierarchical Clustering}

\author{Nikos~Pitsianis}
\affiliation{%
  \department{Electrical~and~Computer~Engineering}
  \institution{Aristotle~University~of~Thessaloniki}
  \city{Thessaloniki} 
  \country{Greece} 
  \postcode{54124}
}
\affiliation{%
  \department{Computer Science}
  \institution{Computer Science, Duke University}
  \city{Durham} 
  \state{NC} 
  \postcode{27710}
  \country{USA}
}
\email{pitsiani@auth.gr}

\author{Dimitris~Floros}
\affiliation{%
  \department{Electrical~and~Computer~Engineering}
  \institution{Aristotle~University~of~Thessaloniki}
  \city{Thessaloniki} 
  \country{Greece} 
  \postcode{54124}
}
\email{fcdimitr@auth.gr}

\author{Alexandros-Stavros~Iliopoulos}
\affiliation{%
  \department{Computer~Science}
  \institution{Duke~University}
  \city{Durham} 
  \state{NC} 
  \postcode{27710}
  \country{USA}
}
\email{ailiop@cs.duke.edu}

\author{Kostas~Mylonakis}
\affiliation{%
  \department{Electrical~and~Computer~Engineering}
  \institution{Aristotle~University~of~Thessaloniki}
  \city{Thessaloniki} 
  \country{Greece} 
  \postcode{54124}
}
\email{mylonakk@auth.gr}

\author{Nikos~Sismanis}
\affiliation{%
  \department{Electrical~and~Computer~Engineering}
  \institution{Aristotle~University~of~Thessaloniki}
  \city{Thessaloniki} 
  \country{Greece} 
  \postcode{54124}
}
\email{nsismani@auth.gr}

\author{Xiaobai~Sun}
\affiliation{%
  \department{Computer~Science}
  \institution{Duke~University}
  \city{Durham} 
  \state{NC} 
  \postcode{27710}
  \country{USA}
}
\email{xiaobai@cs.duke.edu}

\renewcommand{\shortauthors}{N.~Pitsianis et al.}

\begin{abstract}
Calculation of near-neighbor interactions among high dimensional,
irregularly distributed data points is a fundamental task to many
graph-based or kernel-based machine learning algorithms and
applications.  Such calculations, involving large, sparse interaction
matrices, expose the limitation of conventional data-and-computation
reordering techniques for improving space and time locality on modern
computer memory hierarchies.
We introduce a novel method for obtaining a matrix permutation that
renders a desirable sparsity profile.  The method is distinguished by
the guiding principle to obtain a profile that is block-sparse with
dense blocks. Our profile model and measure capture the essential
properties affecting space and time locality, and permit variation in
sparsity profile without imposing a restriction to a fixed pattern.
The second distinction lies in an efficient algorithm for obtaining a
desirable profile, via exploring and exploiting multi-scale cluster
structure hidden in but intrinsic to the data.
The algorithm accomplishes its task with key components for
lower-dimensional embedding with data-specific principal feature axes,
hierarchical data clustering, multi-level matrix compression storage,
and multi-level interaction computations.
We provide experimental results from case studies with two important data
analysis algorithms. The resulting performance is remarkably comparable to the
BLAS performance for the best-case interaction governed by a regularly banded
matrix with the same sparsity.

\end{abstract}

\begin{document}

% document_details.tex

% htt/dl.acm.occs.cfm
%   <concept_id>10010520.10010553.1001056concept_id>
%   <concept_desc>Computer systems organization~Embedded systemconcept_desc>
%   <concept_significance>50concept_significance>
% concept>
%   <concept_id>10010520.10010575.1001075concept_id>
%   <concept_desc>Computer systems organization~Redundancconcept_desc>
%   <concept_significance>30concept_significance>
% concept>
%   <concept_id>10010520.10010553.1001055concept_id>
%   <concept_desc>Computer systems organization~Roboticconcept_desc>
%   <concept_significance>10concept_significance>
% concept>
%   <concept_id>10003033.10003083.1000309concept_id>
%   <concept_desc>Networks~Network reliabilitconcept_desc>
%   <concept_significance>10concept_significance>
% concept>
%ccs2012>  


%%% Local Variables:
%%% mode: latex
%%% TeX-master: "MPDO-main"
%%% End:

% preamble-acm-keywords.tex

%%% Local Variables:
%%% mode: latex
%%% TeX-master: "MPDO-main"
%%% End:

\maketitle

% subfigure annotations
\newcommand{\subfigimg}[4]{%
  \setbox2=\hbox{\includegraphics[width=#1]{#4}}% Store image in box
  \leavevmode\rlap{\usebox2}% Print image
  \rlap{\hspace*{0.3em}\raisebox{\dimexpr\ht2-1em}
  {\footnotesize\textcolor{#3}{{#2}}}}% Print label
  \phantom{\usebox2}% Insert appropriate spacing
}

%%%%%%%%%%%%%%%%%%%%%%%%%%%%%%%%%%%%%%%%%%%%%%%%%%
%%% MAIN BODY

\section{Introduction}
\label{sec:Introduction}
% introduction.tex
% 

We introduce a new method for improving data locality, and thereby
performance on modern computer architectures, for iterative
near-neighbor interaction computations among scattered data points in a
high-dimensional feature space.  Such interactions are a fundamental
computational task in several high-impact and frequently used algorithms
for exploratory data analysis or machine learning applications.
Among others, they arise in the mean-shift (MS)~\cite{fukunaga_estimation_1975,
  Comaniciu2002}, stochastic neighbor embedding
(SNE)~\cite{hinton2003stochastic}, and t-student stochastic neighbor embedding
(t-SNE)~\cite{maaten-hinton2008, VanDerMaaten_accelerating_2014} algorithms.
Iterative interactions are becoming a speed bottleneck as the data sets are
growing bigger.

We give a simple description of iterative interaction computations.
Denote by $\mathcal{S} = \setcond{s_j}{j=1,\ldots,N}$ and
$\mathcal{T} = \setcond{t_{i\phantom{j}}\!\!}{i=1,\ldots,M}$ the source
(or reference) and the target (or response) point set, respectively, in
a $D$-dimensional feature space, $\mathbb{R}^{D}$. The sources may be
the corpus or training data in a stationary setting.  The targets may
correspond to query or test data.  Very often, feature dimensionality is
high and the data sets are big.
For instance, the feature dimension is $49$ with $7 \!\times\! 7$ patch
features, as often used in image analysis methods, and it can be as high
as $\almost 1000$ with GIST descriptors~\cite{oliva2001modeling}.  The
number of data points can be in the order of millions or even billions.

At iteration step $\idxIter$, we denote by
$\mbs{x}_{\idxIter} = [x_{\idxIter}(s_j)]$ the charge (reference) vector
defined on the source set and governed by the charge function $x$.  We
denote by $\mbs{y}_{\idxIter}$ the potential (response) vector at the
target points due to the impact of the charge $\mbs{x}_{\idxIter}$.
In a stationary setting, the near-neighbor interaction computation is
dominated by a matrix-vector multiplication
$\mbs{y}_{\idxIter} = \mat \mbs{x}_{\idxIter}$, where $\mat$ is the
near-neighbor interaction matrix,
\begin{align} 
  \label{eq:interaction-matrix} 
  \alpha_{ij} = f( t_i,  s_j)    
  \quad \text{iff} \quad
  s_j \in \nn(t_i), 
  &&
     1 \leq i \leq M 
\end{align} 
and $f$ is the kernel function governing the source-target interaction.
Near-neighbor relationships are determined by a given distance measure in the
feature space, $\mathbb{R}^D$.  The matrix columns and rows correspond to the
source and target points, respectively.
We are concerned primarily with the non-stationary setting, where one or
both sets may migrate in $\mathbb{R}^{D}$ during the course of iterative
interactions.  While the data coordinates change, the near-neighbor
interaction matrix is updated accordingly, from $\mat_{\idxIter-1}$ to
$\mat_{\idxIter}$.
For convenience, we assume $M=N$ from now on, unless stated otherwise.

The interaction matrix $\mat$ is related to a bipartite graph, with
edges between the source vertex set $\mathcal{S}$ and the target vertex
set $\mathcal{T}$.  Often, the number of neighbors for each target point
is specified to a predetermined, modest parameter value $k \ll N $.  The
matrix profile corresponds to the $k$-nearest neighbors ($k$NN) graph;
it is sparse, and not necessarily symmetric.
Near-neighbor interactions include those in traditional sparse matrix
computations, such as Laplacian diffusion in a low-dimensional
space-time coordinate space, over points on a regular or irregular mesh.

Near-neighbor interactions among large data points in a high-dimensional
feature space tend to suffer from the widening gap between processor
speed and memory speed~\cite{rixner_memory_2000}.  Furthermore, they
expose the limitations of conventional techniques for improving data
locality via data-computation reordering, at the system level and/or at
the application algorithm level.
Locality improvement at the system level relies on a particular program
instantiation of an application algorithm~\cite{Mellor-Crummey2001}.  At
the application algorithm level, it has a larger scope to alter program
instantiation and memory access footprint~\cite{george_solution_1986,
  barnard1995, Mellor-Crummey2001}; the latter, however, is challenged
by high dimensionality.

When the feature dimension $D$ is high, then $N \ll 2^{D}$.  This
relationship between data size and dimension speaks plainly to the fact
that the data points are sparsely distributed in the feature space,
unlike the points on, and well-connected by, a low-dimensional mesh.
By conventional matrix ordering techniques, as we shall show shortly,
the near-neighbor interaction matrix remains highly sparse within an
optimal envelope; for conventional sparsity profile types,
see~\cite{Mellor-Crummey2001, barnard1995} and the references therein.
By the minimum bandwidth envelope, for example, the bandwidth in a row
is typically much larger than the number of near neighbors, i.e., the
number of non-zero elements in the row.  This phenomenon may be viewed
as a shadow of the so-called curse of dimensionality.  We introduce our
method to step away from this shadow.

We explore the potential, at the algorithm level, to further utilize
processor speed and circumvent memory access latency due to scattered
referencing among high-dimensional irregular data, by means of exploring
and exploiting multi-scale cluster structure that is hidden in but
intrinsic to the data set.  The method itself is data adaptive.
There are two main aspects to the novelty of our approach.

The first novelty lies in the principled pursuit of a matrix profile
that is \emph{block-sparse with dense blocks}.  The guiding principle
suggests that neighboring, interacting elements should be clustered as
densely as possible.  In an ideal ordering, the sparsity profile has at
least a two-level structure: very sparse in nonzero sub-matrices
(blocks); much denser within in each nonzero block.
We present in Sections~\ref{sec:guiding-principle}
and~\ref{sec:matrix-ordering-evaluation} a descriptive measure of what
we term the patch density of a matrix, for quantitative evaluation of
and comparison between particular profiles.  The profile model and
measure reflect and capture essential properties affecting space and
time locality for sparse matrix-vector multiplications.

The second contribution lies in a new algorithm for efficiently locating
an ordering that renders a near-optimal matrix profile, based on
intrinsic multi-scale relationships among the data.
The algorithm is composed of key components for lower-dimensional
embedding with data-specific principal feature axes, hierarchical data
clustering, multi-level matrix compression storage, and multi-level
interaction computations. We describe the algorithm in
Section~\ref{sec:matrix-ordering-algorithm}.

We assess the efficacy of our method with two case studies, involving the
applications of the MS and \tsne\ algorithms to large, high-dimensional data
sets. We describe the case studies briefly in
Section~\ref{sec:CaseStudies}. Experimental results are presented in
Section~\ref{sec:Experiments}.
Comparisons between different ordering schemes are made with respect to
patch density measure and throughput performance, besides visual
inspection and comparison of the sparsity profiles.  The results
demonstrate
\begin{inparaenum}[(i)]
\item superior performance to that of state-of-the-art ordering methods
  at the algorithm level; and
\item remarkable consistency between visual profiles and patch density
  measure on the one side and execution time and throughput on the
  other.
\end{inparaenum}
We conclude the paper with a brief overview additional remarks and a brief
overview of related work in Section~\ref{sec:related-work-discussion}.

%%% Local Variables:
%%% mode: latex
%%% TeX-master: "MPDO-main"
%%% End:

%  LocalWords:  datasets

\section{Maximum patch-density ordering}
\label{sec:matrix-ordering-intro}
% matrix-ordering-intro.tex
% 

In this section, we first introduce our guiding principle for
identifying matrix sparsity profiles that are conducive to better space
and time locality for near-neighbor interactions among data points in a
high-dimensional space.
We materialize the principle into a quantitative measure of the sparse
profile of any given ordering and its maximal value among all orderings.
We term this measure the patch density score.  An optimal matrix
ordering renders the maximal patch density score.
Then, we present an efficient and effective algorithm for attaining a
near-optimal matrix ordering.

%%% Local Variables:
%%% mode: latex
%%% TeX-master: "MPDO-main.tex"
%%% End:

\subsection{Block-sparse with dense blocks}
\label{sec:guiding-principle}
% matrix-ordering-principle.tex

\begin{figure}
  \centering
  % 
  % ---------- block antidiagonal
  % 
  \begin{subfigure}{0.24\linewidth}
    \centering
    \captionsetup{width=0.7\linewidth}%
    \includegraphics[width=\linewidth]{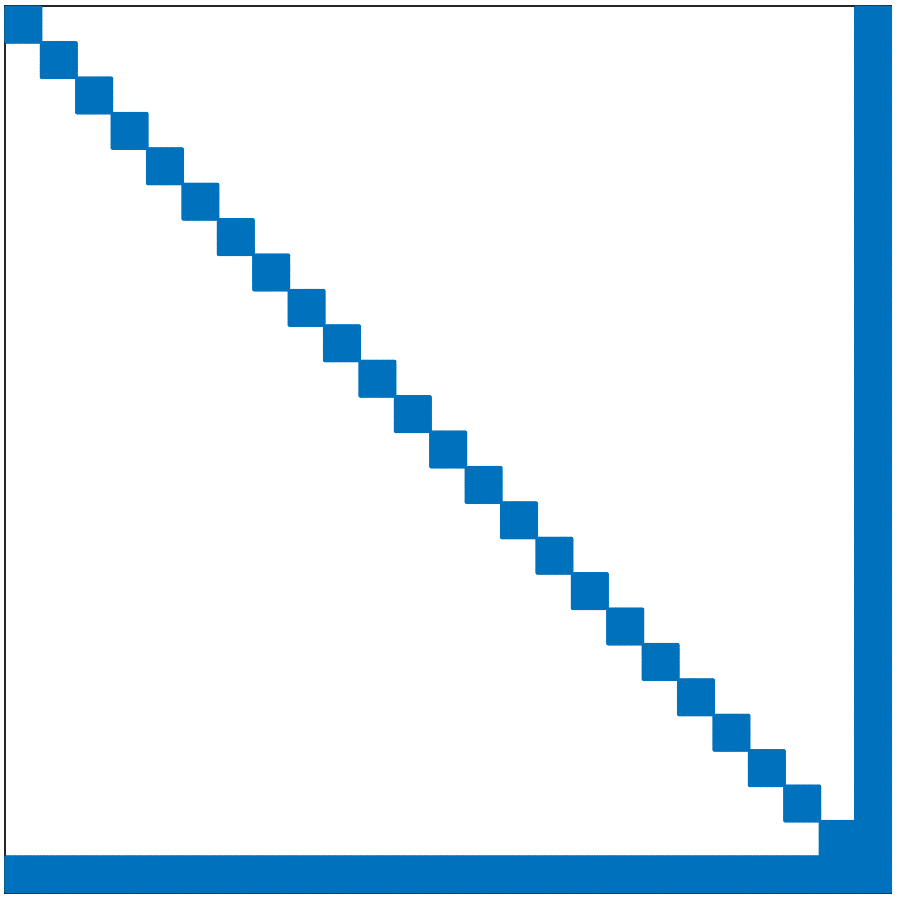}
    \caption{%
      $\patchDensity \approx 0.20$\\
      \phantom{\bf(a) }$\patchDensityKern = 21.0$}
    \label{fig:example-profile-base}
  \end{subfigure}%
  \hspace*{\fill}
  % 
  % ---------- block permutation
  % 
  \begin{subfigure}{0.24\linewidth}
    \centering
    \captionsetup{width=0.7\linewidth}%
    \includegraphics[width=\linewidth]{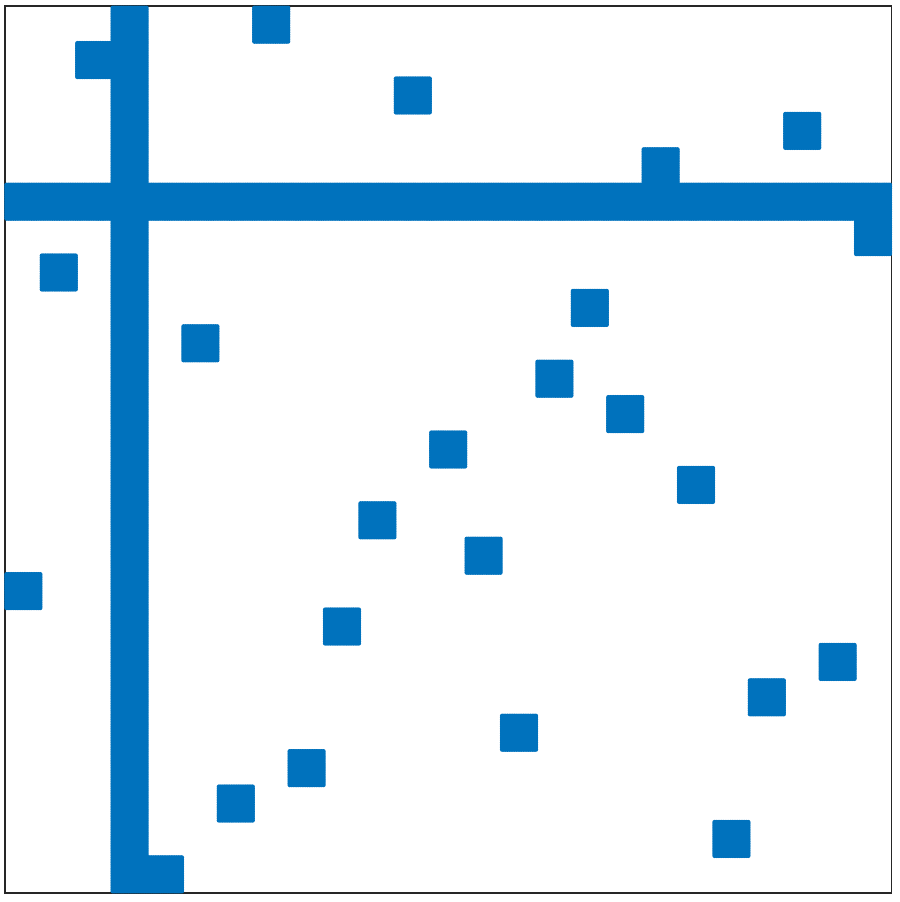}
    \caption{%
      $\patchDensity \approx 0.20$\\
      \phantom{\bf(b) }$\patchDensityKern = 20.9$}
    \label{fig:example-profile-permute-block}
  \end{subfigure}%
  \hspace*{\fill}
  % 
  % ---------- row permutation
  % 
  \begin{subfigure}{0.24\linewidth}
    \centering
    \captionsetup{width=0.7\linewidth}%
    \includegraphics[width=\linewidth]{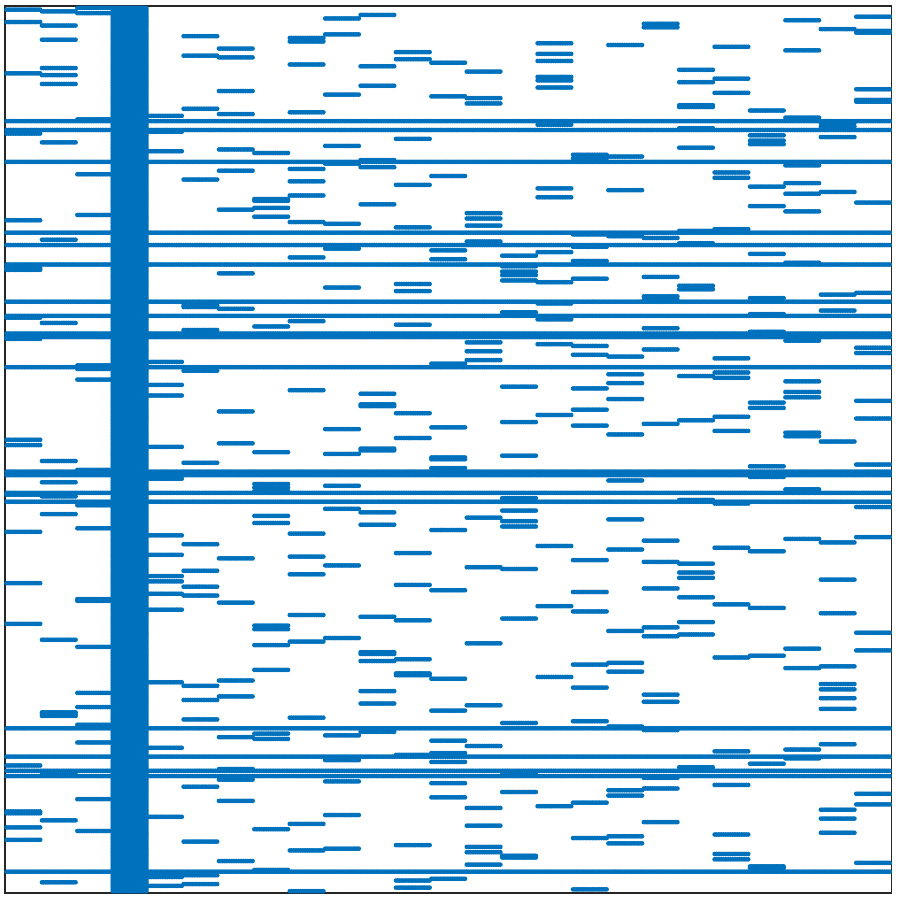}
    \caption{%
      $\patchDensity \ge 0.05$\\
      \phantom{\bf(c) }$\smsp\patchDensityKern = 10.5$}
      \label{fig:example-profile-permute-block-row}
  \end{subfigure}%
  \hspace*{\fill}
  % 
  % ---------- row-column permutation
  % 
  \begin{subfigure}{0.24\linewidth}
    \centering
    \captionsetup{width=0.7\linewidth}%
    \includegraphics[width=\linewidth]{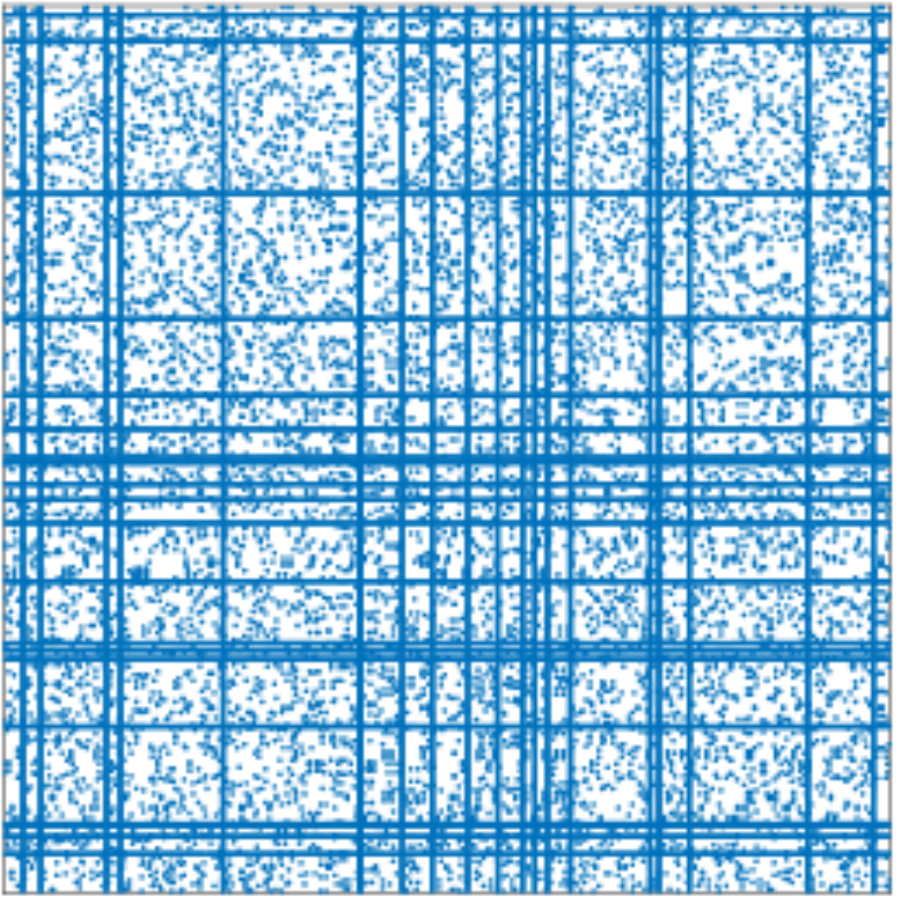}
    \caption{%
      $\patchDensity \ge 0.01$\\
      \phantom{\bf(d) }$\smsp\patchDensityKern = 4.5$}
    \label{fig:example-profile-permute-block-row-col}
  \end{subfigure}%
  \caption{%
    Sparsity profiles of the same $500 \times 500$ matrix in four different
    orderings: %
    \emph{(a)}~Block arrowhead with full $20 \times 20$ blocks; %
    \emph{(b)}~obtained from (a) with a random permutation among the block
    rows and columns; %
    \emph{(c)}~obtained from (b) with a random permutation among the rows; and %
    \emph{(d)}~obtained from (c) with a random permutation among the
    columns. %
    Patch density is maximal for (a) and (b), the best case.  It is reduced
    for (c), and further dropped for (d), the base case.
    The proposed patch density measure $\patchDensity$ and numerical density
    estimate $\patchDensityKern$, defined in~\eqref{eq:patch-density-definition}
    and~\eqref{eq:kernel-patch-density}, respectively, are shown for each
    ordering.  The $\patchDensityKern$-scores (evaluated using $\sigma = 10$)
    correlate well with the $\patchDensity$-scores. %
  }
  \label{fig:example-profiles} 
\end{figure}

Given a sparse matrix, we seek row and column re-orderings such that the
sparsity profile is block-sparse with dense blocks.
We illustrate first in Fig.~\ref{fig:example-profiles} four sparsity
profiles of the same matrix as per four different orderings; their
relationship to one another is explained in the caption.
After specifying what we mean by \textit{block sparsity} and
\textit{dense blocks}, we argue that the two properties are integral to
each other, and that they are essential to relating data locality to
matrix sparsity profile and further characterizing this relationship.

% FIGURE: example sparsity profiles for block-arrowhead-like interactions
% figure-example-profiles.tex

%%% Local Variables:
%%% mode: latex
%%% TeX-master: "MPDO-main"
%%% End:

In what follows, we denote by $\nnz{\mat}$ the total number of nonzero
elements in a near-neighbor interaction matrix $\mat$.  With $k$NN
interaction in particular, $\nnz{\mat} = k N$.  Denote by $\area(\mat)$
the product of the number of rows and the number of columns, i.e., the
total number of elements in $\mat$.

\paragraph{Dense blocks}

Assume that the matrix in a certain ordering has a relatively denser
sub-matrix, or simply block $\submat$, such as those in
Figs.~\ref{fig:example-profile-base}
and~\ref{fig:example-profile-permute-block}.
By ``relatively denser,'' we mean that the ratio
$\nnz{\submat} / \!\area(\submat)$ is significantly higher than
$\nnz{\mat} / \!\area(\mat)$.
When $\mat$ corresponds to a bipartite graph between the entire source
and target data sets, block $\submat$ corresponds to a bipartite
sub-graph with a cluster of target points sharing many of their
neighbors in a cluster of source points.  The density of $\submat$ may
get close to $1$, despite the sparsity of $\mat$.
For matrix-vector multiplication, such a dense block implies good
locality in reading the charge sub-vector over the source cluster as
well as good locality in writing to the response sub-vector over the
target cluster, provided that the charge and response vectors are placed
in memory according to the cluster structure.

\paragraph{Block sparsity}

The nonzero blocks need not have the same size.  By ``block-sparse,'' we mean
that the dense blocks are sparsely distributed, i.e., the number of blocks is
far smaller than $\nnz{\mat}$.
In the trivial and extremely degenerate case, there are exactly
$\nnz{\mat}$ nonzero blocks: a $1 \!\times\! 1$ block for each nonzero
element.  Next to this trivial extreme is the case where there are
$\nnz{\mat}/c$ small blocks scattered throughout the matrix, for some
small number $c$ which reflects the average size of the nonzero but
small blocks. The block-sparsity condition discourages such degenerate
cases.

\paragraph{Discussion}

The principled sparsity profile model has two important consequences.
The first one may be subtle.  The model admits variations in matrix
profiles, without imposing a fixed profile pattern.  For instance, the
profiles in Figs.~\ref{fig:example-profile-base}
and~\ref{fig:example-profile-permute-block} are considered equivalent in
principle.  This principled equivalence in sparsity profiles
\emph{unifies} several previously existing profile patterns.
Particularly, the block-sparse profile of an arrowhead shape in
Fig.~\ref{fig:example-profile-base} is in principle as good as its
banded counterpart with the same number of nonzero elements.

Secondly, our principled sparsity profile is at the level of global
matrix reordering. It applies naturally to the profiles of blocks, and
hence leads to a \emph{hierarchical} sparsity profile.

These properties make the model adaptive in practical computation to
architecture specific performance tuning and/or to application specific
customization in order to exploit additional structure.

%%% Local Variables:
%%% mode: latex
%%% TeX-master: "MPDO-main.tex"
%%% End:

%  LocalWords:  reorderings

\subsection{Profile measure and optimal ordering}
\label{sec:matrix-ordering-evaluation}
% matrix-ordering-measure.tex 

We have established a concrete measure of the sparsity profile of a matrix
$\mat$ in any particular ordering.  The measure favors sparsity profiles that
match the block-sparse with dense blocks principle.  For convenience in
description, we introduce the concept of a patch covering of the nonzero
elements in $\mat$.  A patch covering is a set of non-overlapping blocks
(patches) $\setenum{\patch}$ such that every nonzero element of $\mat$ lies
within a patch in the covering.  The covering size is
$ \setcard{\setenum{\patch}}$, the number of patches in $ \setenum{\patch} $.
The covering area is
$ \area( \setenum{\patch} ) = \sum_{\idxPatch = 1}^{\setcard{\setenum{\patch}}}
\area(\patch) $.
The ratio $\nnz{\mat} / \!\area(\setenum{\patch}) $ is the average density of the
nonzero elements over the covering area.
Let $\patchCovering(\mat)$ be the set of all possible patch coverings of $\mat$
in a particular ordering.
Taking into account the block-sparse condition, which entails that the
covering size be made as small as possible, we define the patch
(covering) density measure as
\begin{equation}
  \label{eq:patch-density-definition}
  \patchDensity\left( \mat \right) = 
  \max_{\setenum{\patch} \in \patchCovering(\mat)}
  \frac{ 1 }{ \setcard{\setenum{\patch}} }
  \frac{ \nnz{\mat} }{ \area(\setenum{\patch}) } .
\end{equation}
The best patch covering reaches the measure $\beta(\mat)$, which is
specific to any particular matrix ordering in rows and columns.
We now define the optimal matrix row-column orderings as
\begin{equation}
  \label{eq:maximum-patch-ordering}
  \left( \permTgt, \permSrc \right)^{*}  = 
  \argmax_{(\permTgt, \permSrc)}
  \patchDensity\left( \mat(\permTgt, \permSrc )  \right) ,
\end{equation}
where $\permSrc$ and $\permTgt$ are permutations among columns (sources) and
rows (targets), respectively.

Following the principle elaborated in
Section~\ref{sec:guiding-principle}, the patch density
measure~\eqref{eq:patch-density-definition} relates to several existing
profile patterns and measures used in sparse matrix computation.
Limited by the manuscript length, we provide only a very simplified
connection.  With the same matrix size and number of nonzero elements,
and the same number of dense blocks of equal size, block sparsity
profiles with an arrowhead pattern, banded pattern, or any other
pattern, reach to the same and maximal patch density score.  This
principled equivalence among such patterns amounts to a unification of
them at the level of dense blocks, which correspond to cluster-cluster
interactions and can be translated to space and time locality during
matrix-vector multiplications and other operations.

The patch density score of \eqref{eq:patch-density-definition} may be
normalized by, say, the score for a block-tridiagonal matrix with the
same size and number of nonzero elements.  This is not necessary,
however, because such normalization does not alter the optimization of
\eqref{eq:maximum-patch-ordering}.

%%% Local Variables:
%%% mode: latex
%%% TeX-master: "MPDO-main.tex"
%%% End:

%  LocalWords:  sigmoid tridiagonal

\subsection{Numerical measure estimate}
\label{sec:matrix-ordering-estimate}
% matrix-ordering-estimate.tex
% 

The patch (covering) density measure $\patchDensity(\mat)$
of~\eqref{eq:patch-density-definition} is in a combinatorial expression.
Computation of the ordering-specific measure itself is NP-hard, let alone the
search by~\eqref{eq:maximum-patch-ordering} for the ordering with maximal patch
density. We present in this section a relaxed and differential expression to get
a numerical estimate of the $\beta$-score.

A relaxation shall be based on our profile principle as well as on the
relationship~\eqref{eq:interaction-matrix} between the placement of
nonzero elements in the matrix and the underlying near-neighbor
relationships among the data points.
Denote by $ \nonzeros(\mat)$ the set of indices for the nonzero elements
of $\mat$, $ \nonzeros(\mat) = \setcond{ (i,j) }{ \alpha_{ij} \ne 0 }$.
This set represents the placement of the nonzero elements in the matrix
by a particular ordering.
Among many possible ways to relax the $\patchDensity$-score, we use the
following $\patchDensityKern$-score,
\begin{equation}
  \label{eq:kernel-patch-density}
  \patchDensityKern \left( \mat; \scale \right) = 
  \frac{1}{\scale \nnz{\mat}} 
  \sum_{ \idxA, \idxB \in \nonzeros(\mat)   } 
  \exp \left\{ - \frac{\norm{ \idxA - \idxB }_2^2 }{ \scale^2 } \right\}, 
\end{equation}
where $\sigma$ is a scale parameter.  The Gaussian function is defined over
$ \nonzeros(\mat)\times \nonzeros(\mat) $.
A peak in the Gaussian function corresponds to a dense block in the matrix,
where the block size is regulated by $\sigma$. All dense blocks do not
necessarily have the same size, depending on whether the data points are
clustered densely or loosely.  In other words, the essential properties of the
best patch covering in the combinatorial description are captured by the
Gaussian function with smooth connections between interacting points and
interacting point clusters.

\begin{figure*}
\begin{subfigure}{0.03\linewidth}
    \centering
    % 
    % sift
    % 
    \vspace*{-1.5em}
    \rotatebox[origin=c]{90}{\textbf{SIFT}}
    \\[16em]
    % 
    % gist
    % 
    \rotatebox[origin=c]{90}{\textbf{GIST}}
  \end{subfigure}
  % 
  % 
  % ----- RANDOM
  % 
  \begin{subfigure}{0.150\linewidth}
    \centering
    % 
    % sift
    % 
    \includegraphics[width=\linewidth]{%
      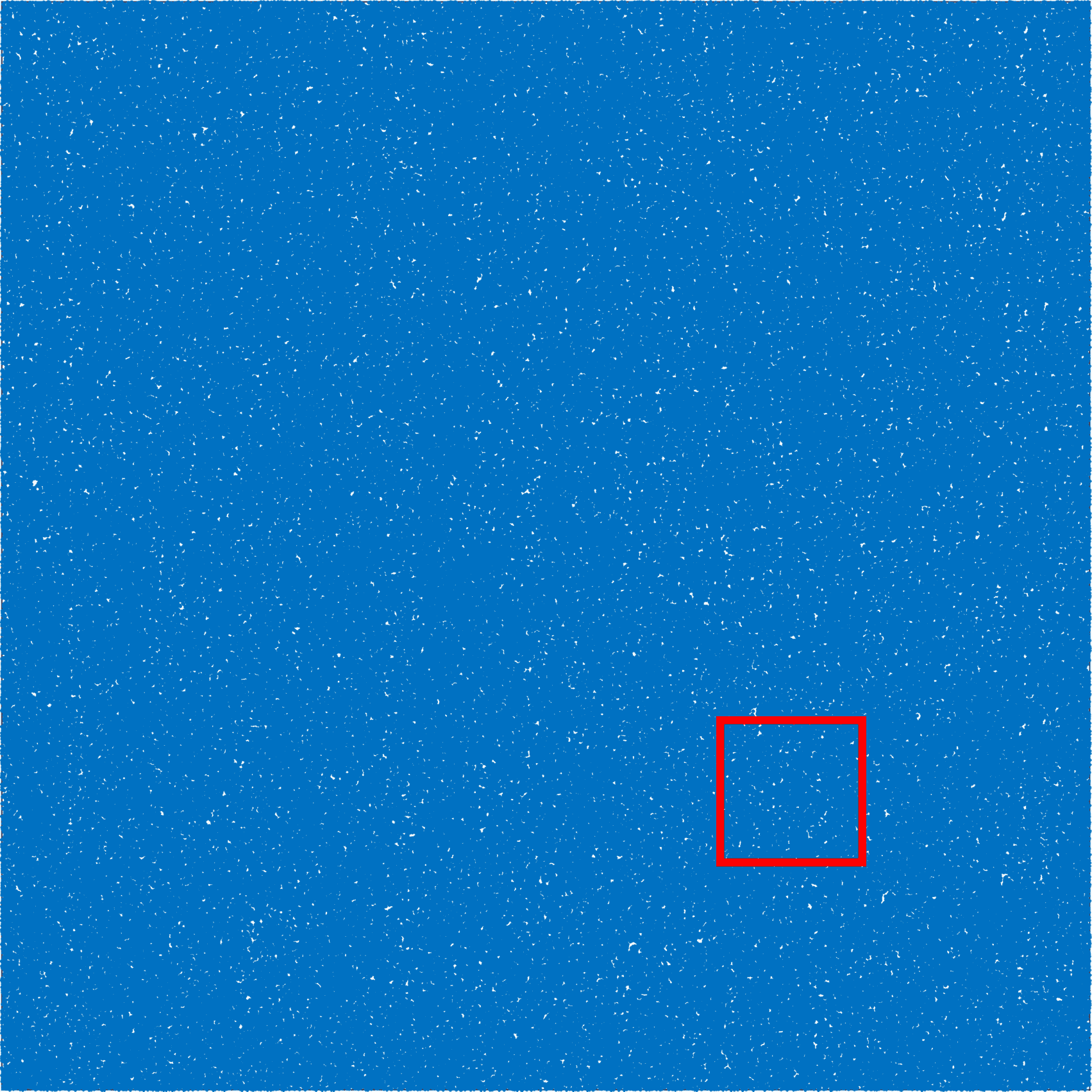%
    }
    \\
    % 
    % sift roi
    % 
    \includegraphics[width=\linewidth]{%
      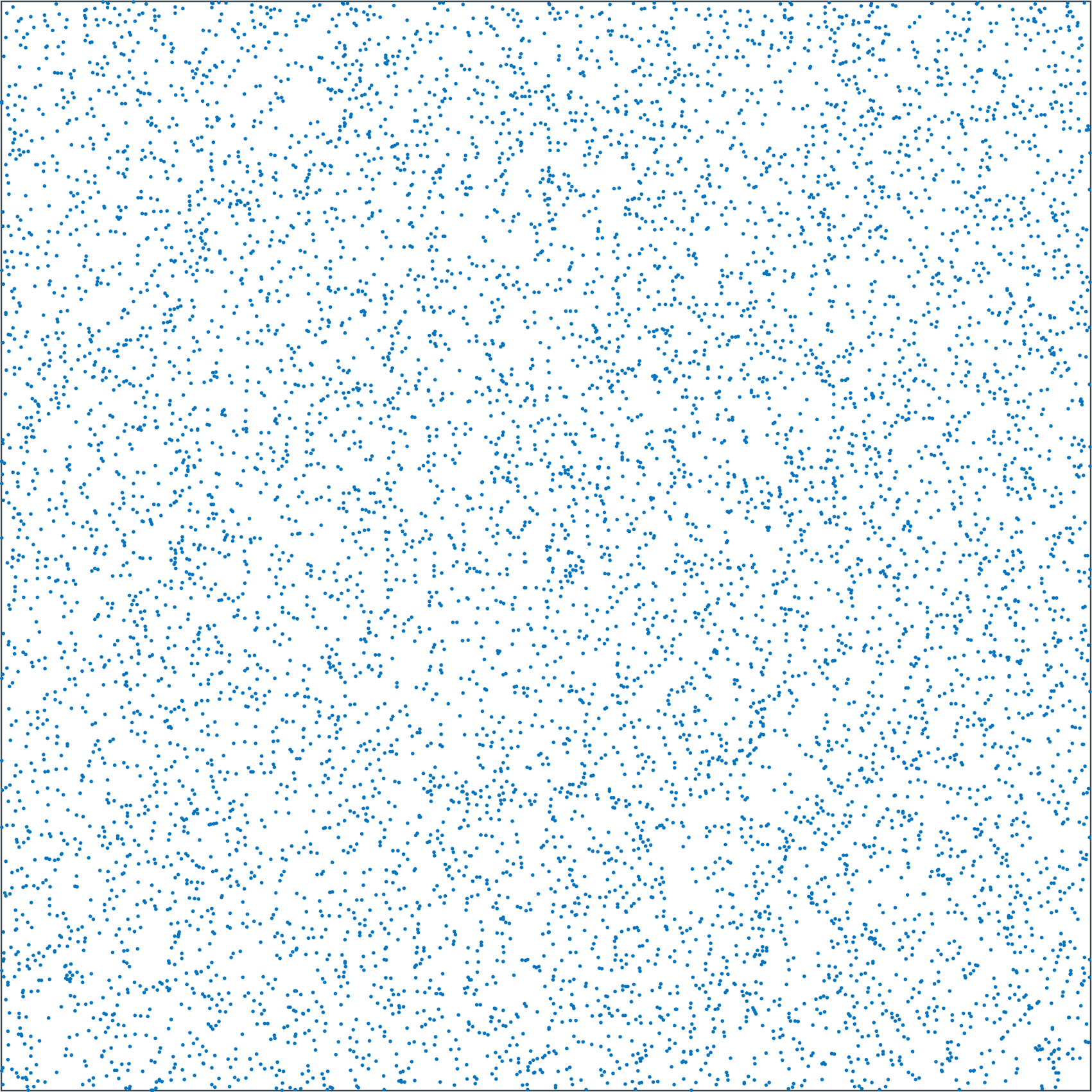%
    }
    \\[1em]
    % 
    % gist
    % 
    \includegraphics[width=\linewidth]{%
      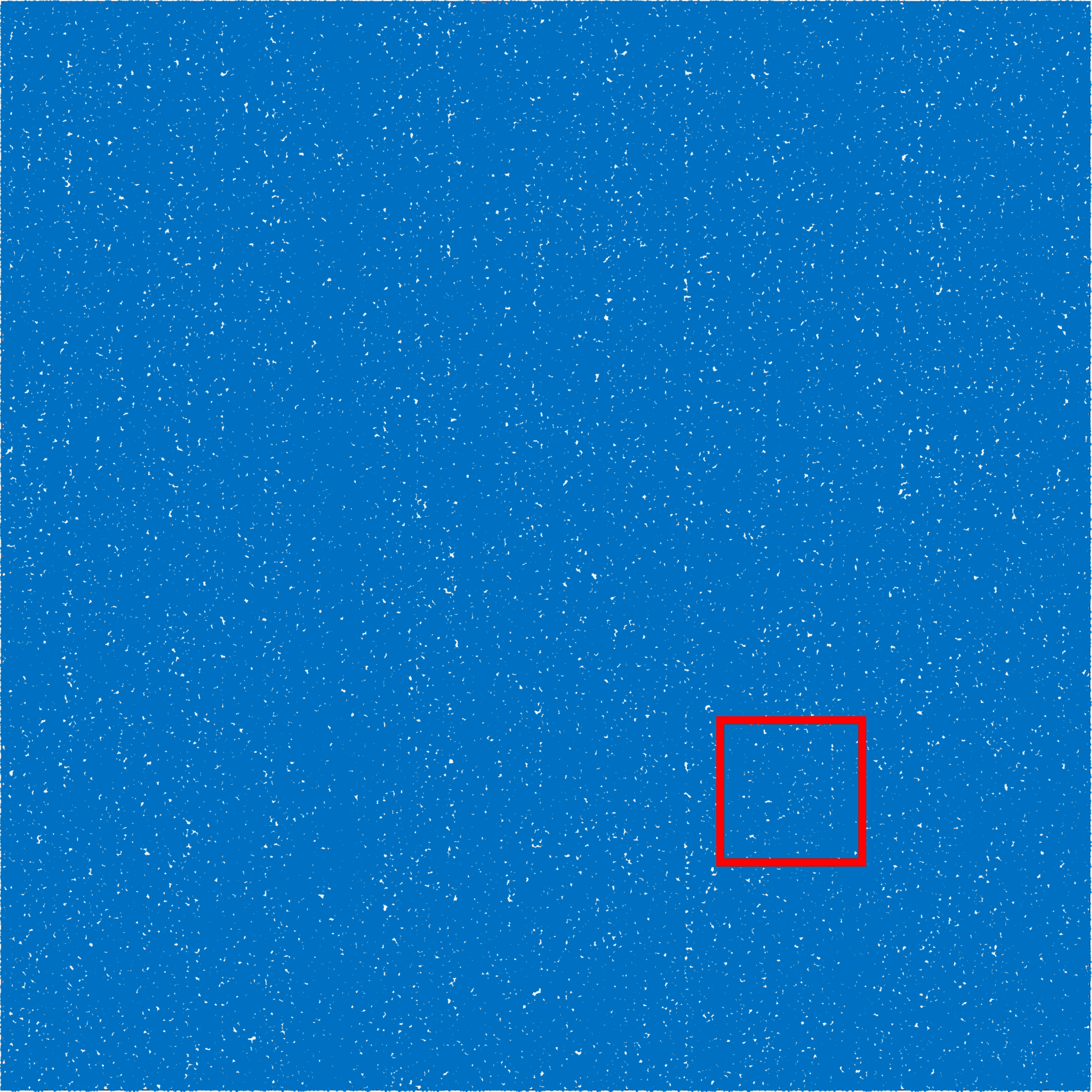%
    }
    \\
    % 
    % gist roi
    % 
    \includegraphics[width=\linewidth]{%
      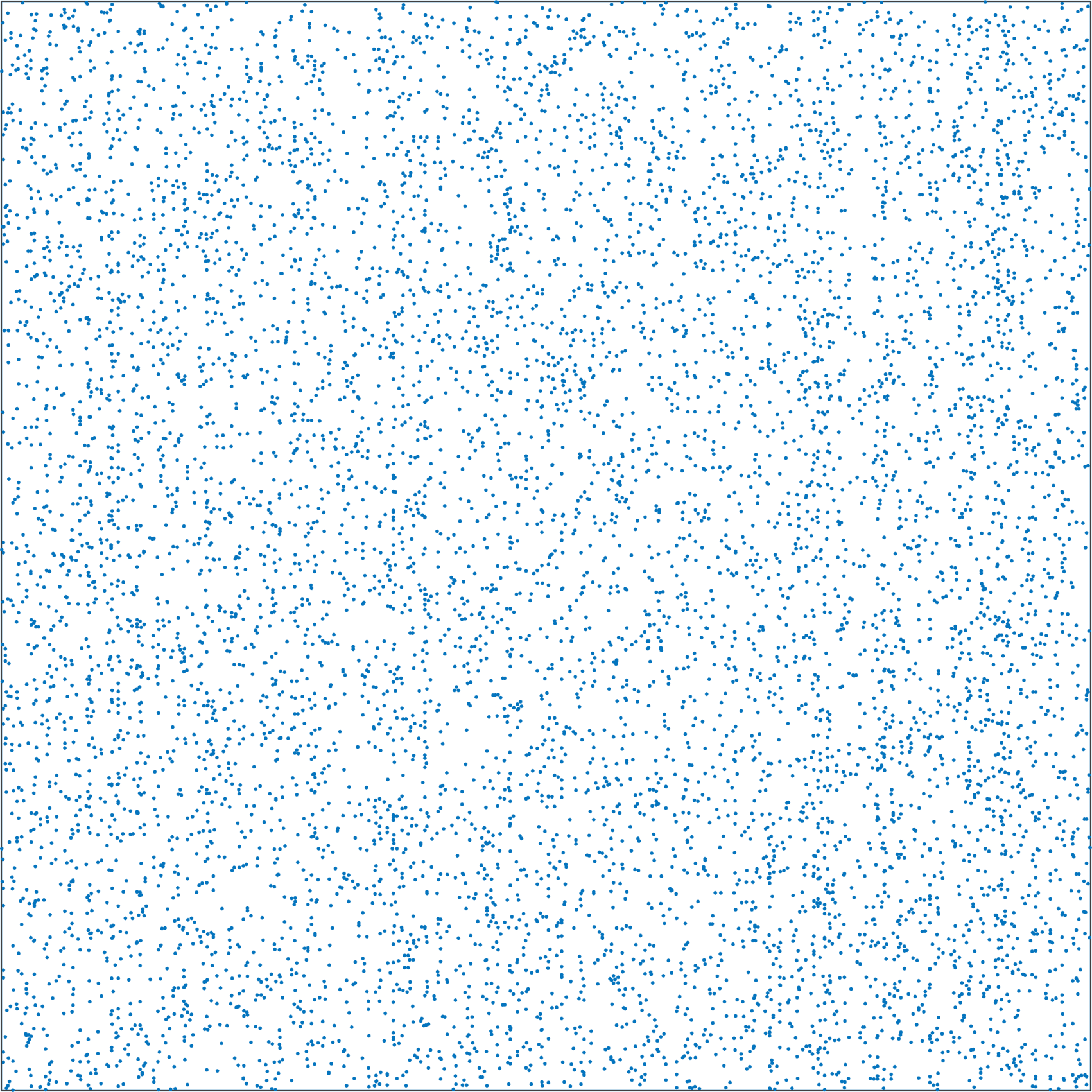%
    }
    \caption{scattered}
  \end{subfigure}
  \hspace*{\fill}
  % 
  % 
  % ----- RCM
  % 
  \begin{subfigure}{0.150\linewidth}
    \centering
    % 
    % sift
    % 
    \includegraphics[width=\linewidth]{%
      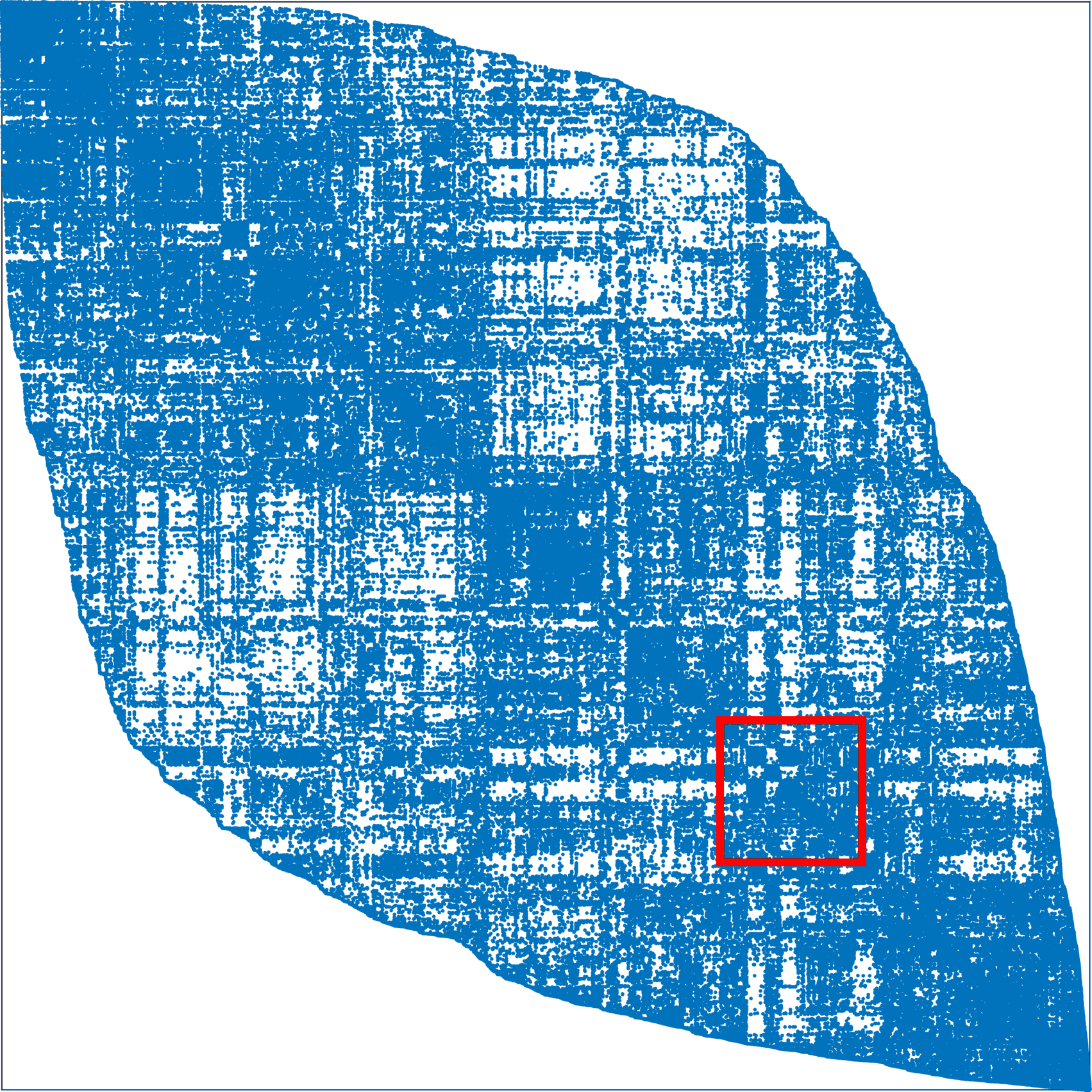%
    }
    \\
    % 
    % sift roi
    % 
    \includegraphics[width=\linewidth]{%
      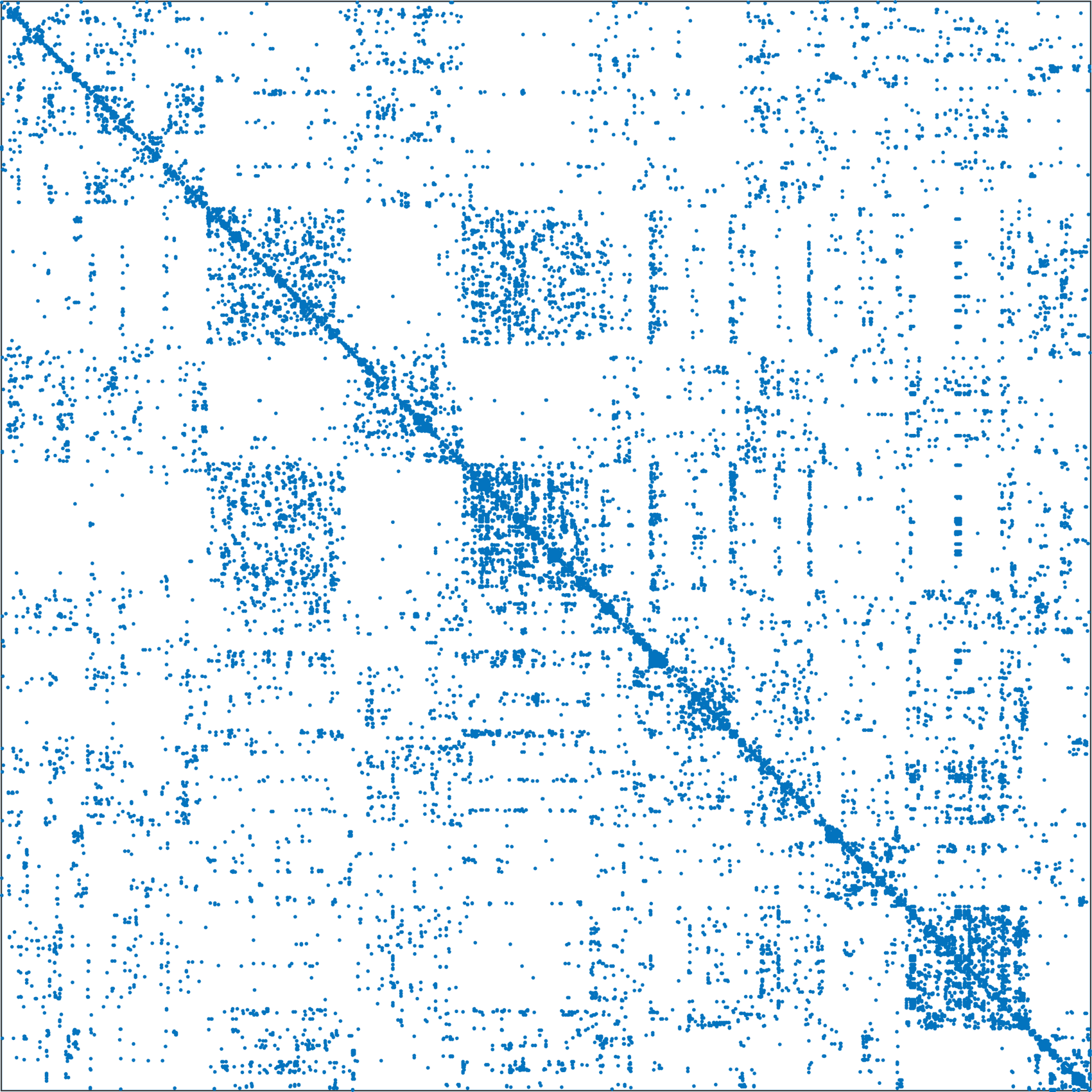%
    }
    \\[1em]
    % 
    % gist
    % 
    \includegraphics[width=\linewidth]{%
      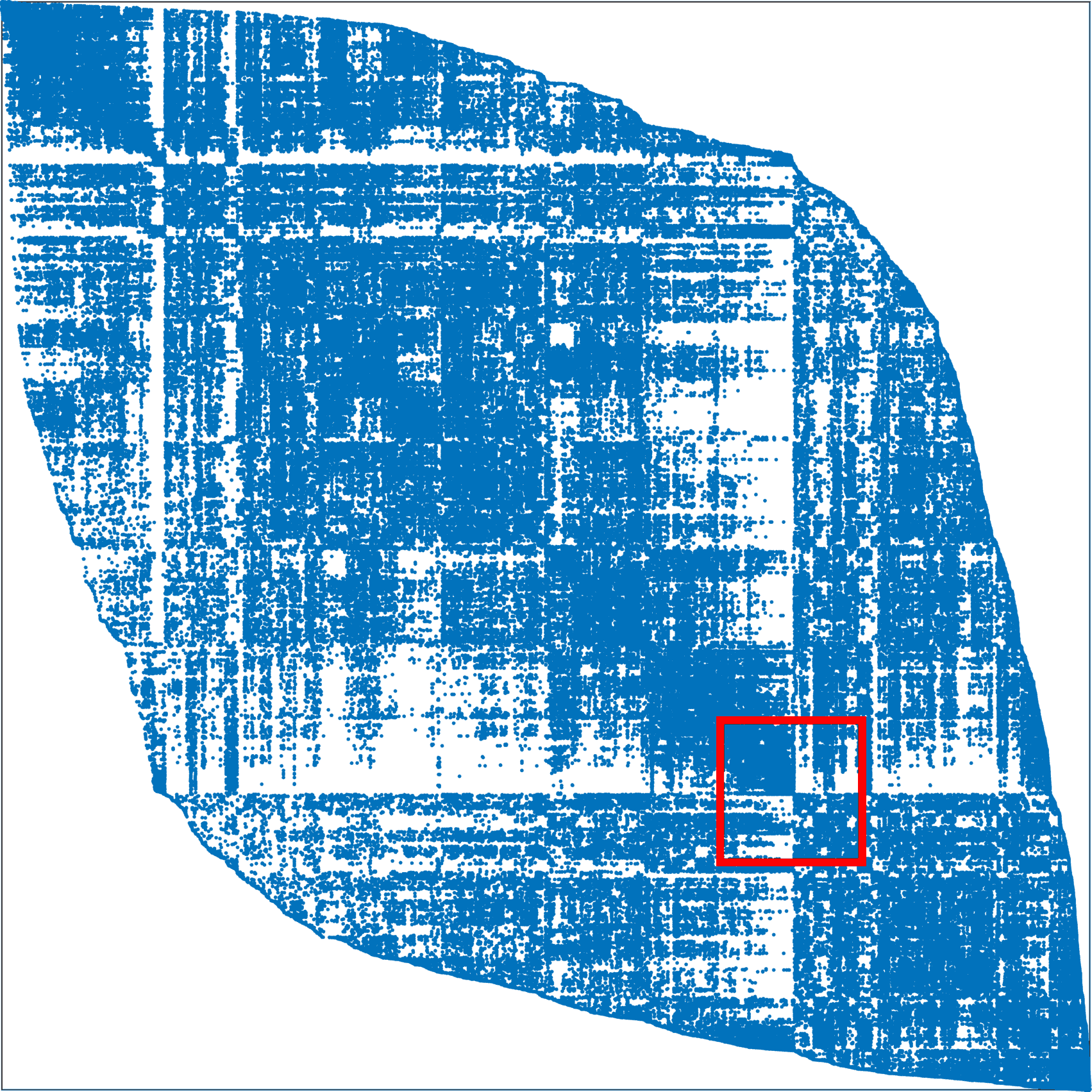%
    }
    \\
    % 
    % gist roi
    % 
    \includegraphics[width=\linewidth]{%
      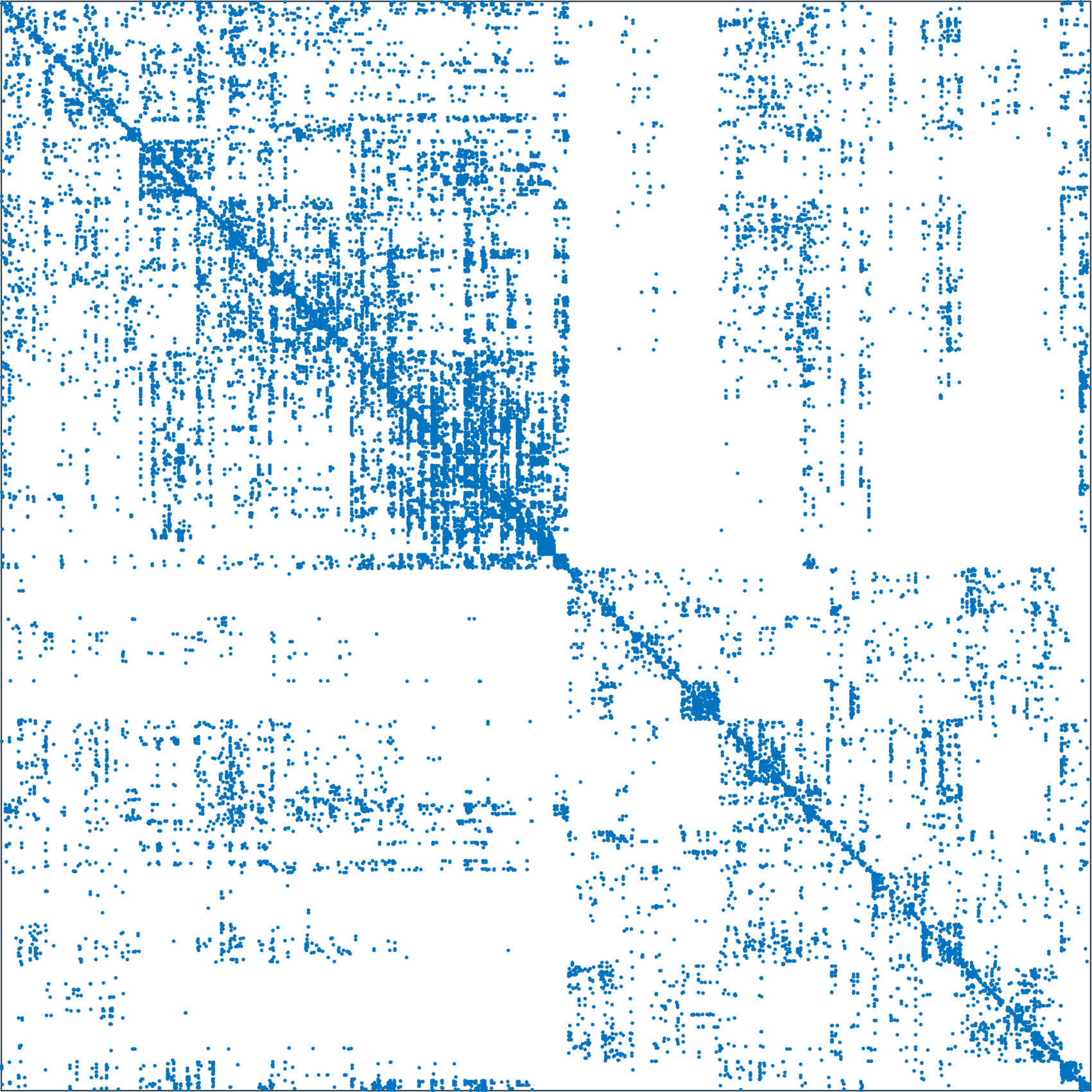%
    }
    \caption{rCM}
  \end{subfigure}
  \hspace*{\fill}
  % 
  % 
  % ----- 1D
  % 
  \begin{subfigure}{0.150\linewidth}
    \centering
    % 
    % sift
    % 
    \includegraphics[width=\linewidth]{%
      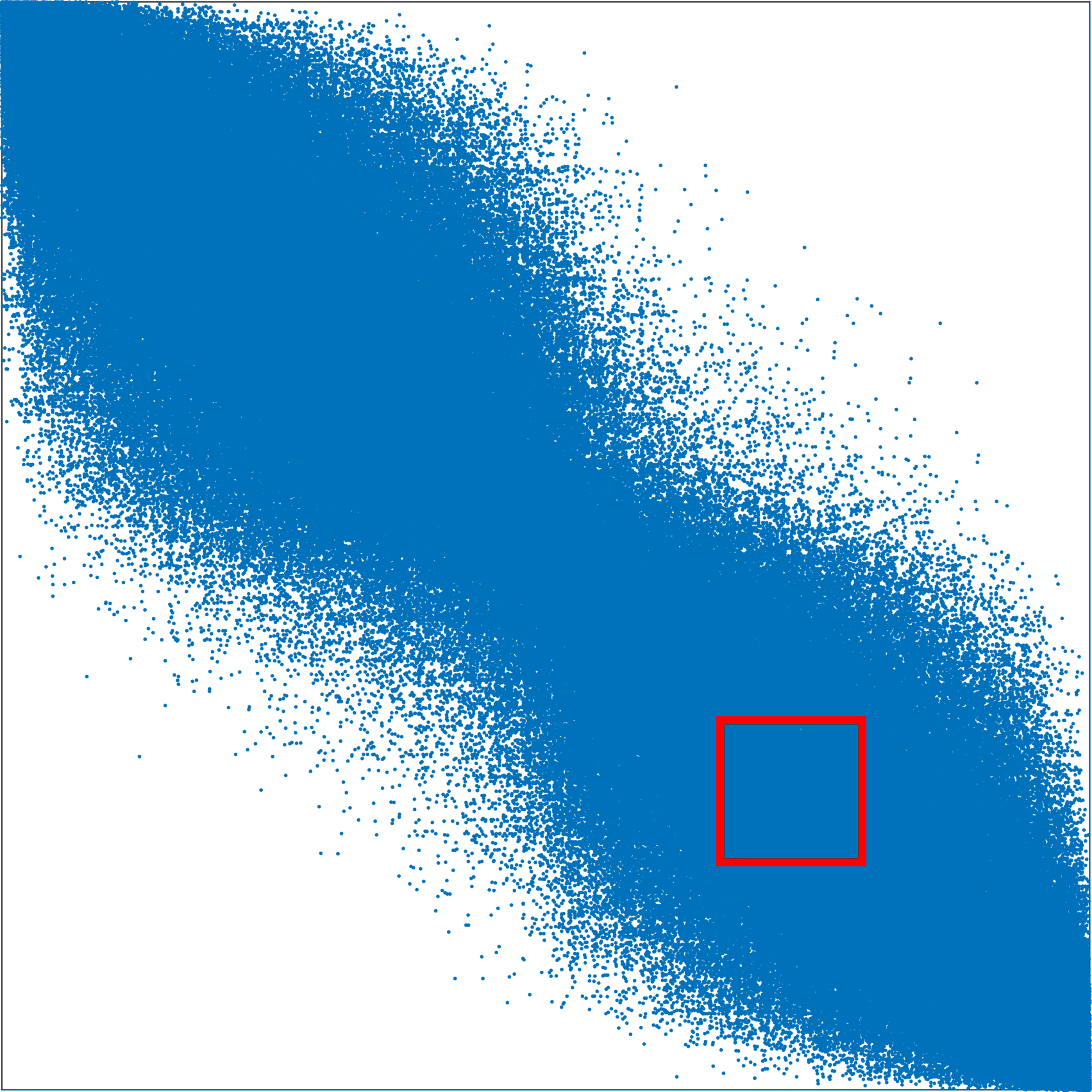%
    }
    \\
    % 
    % sift roi
    % 
    \includegraphics[width=\linewidth]{%
      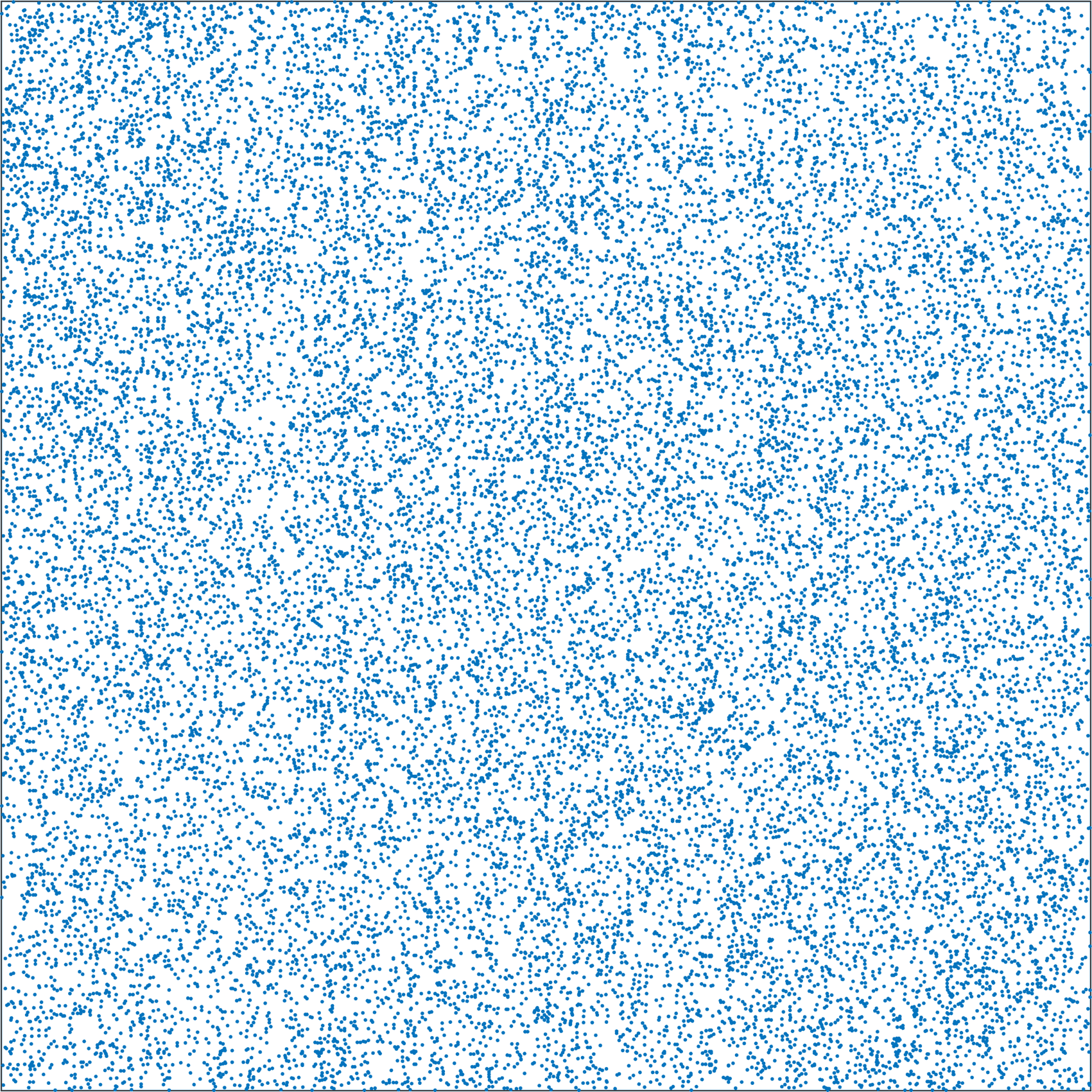%
    }
    \\[1em]
    % 
    % gist
    % 
    \includegraphics[width=\linewidth]{%
      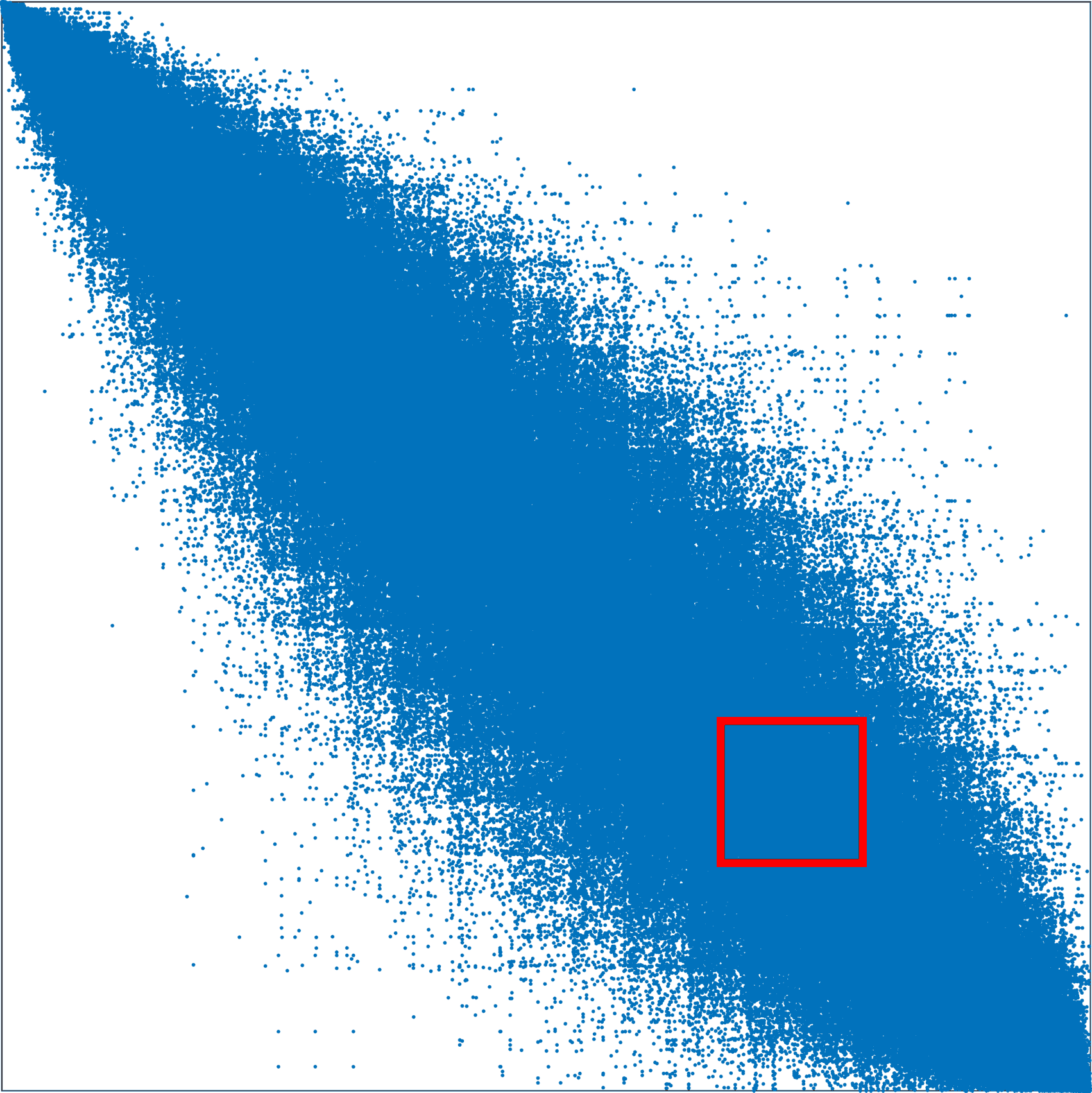%
    }
    \\
    % 
    % gist roi
    % 
    \includegraphics[width=\linewidth]{%
      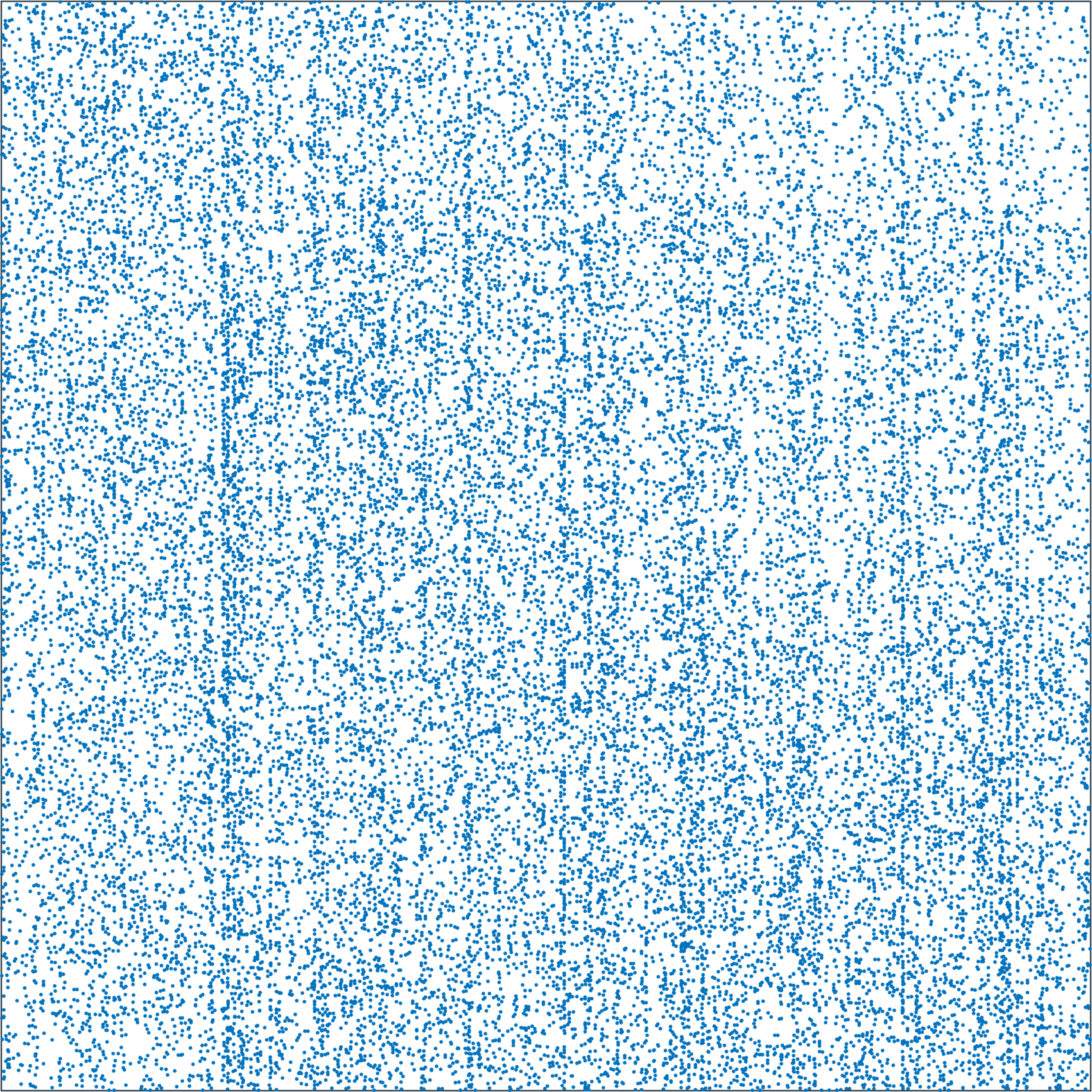%
    }
    \caption{1D}
  \end{subfigure}
  \hspace*{\fill}
  % 
  % 
  % ----- LEXICOGRAPHIC 2D
  % 
  \begin{subfigure}{0.150\linewidth}
    \centering
    % 
    % sift
    % 
    \includegraphics[width=\linewidth]{%
      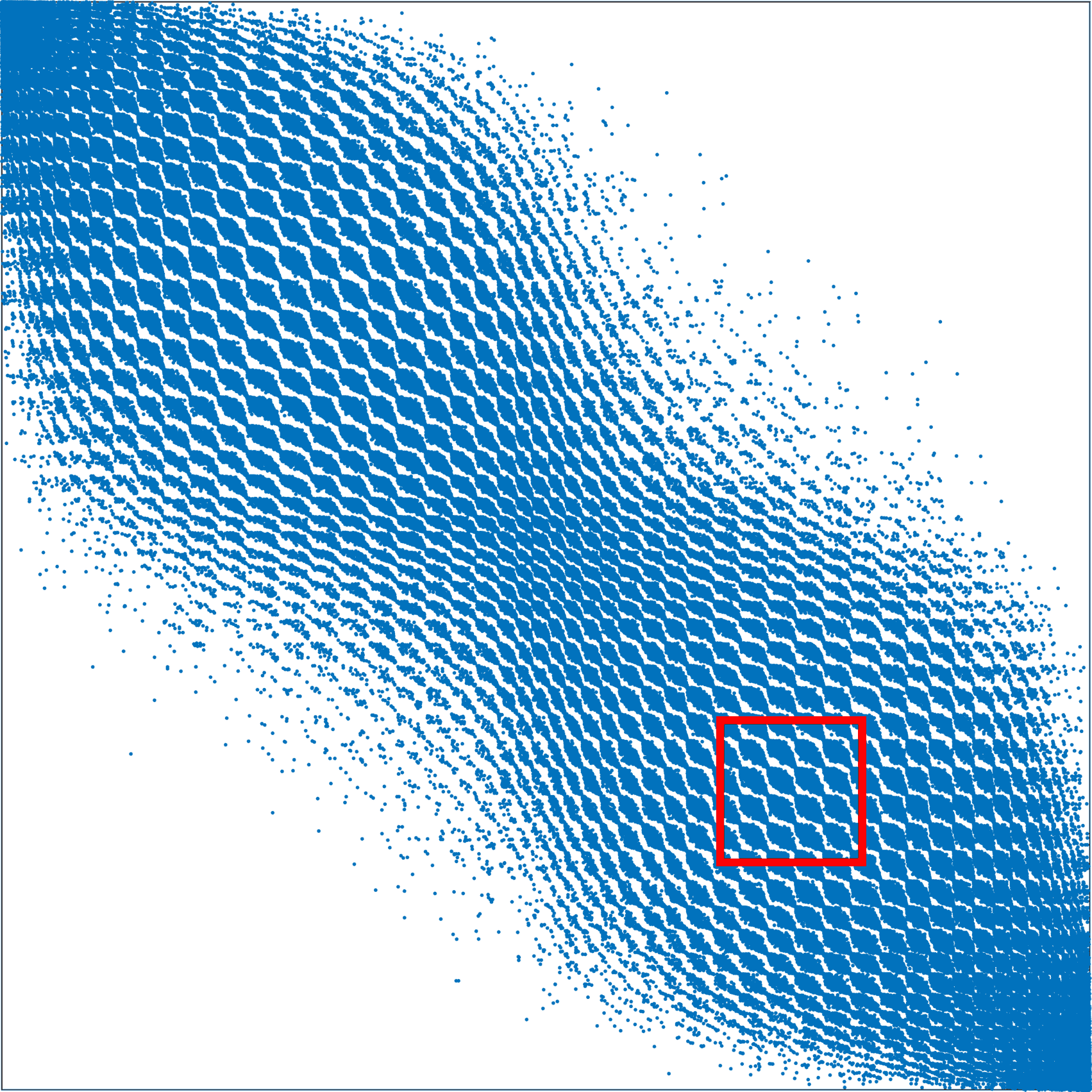%
    }
    \\
    % 
    % sift roi
    % 
    \includegraphics[width=\linewidth]{%
      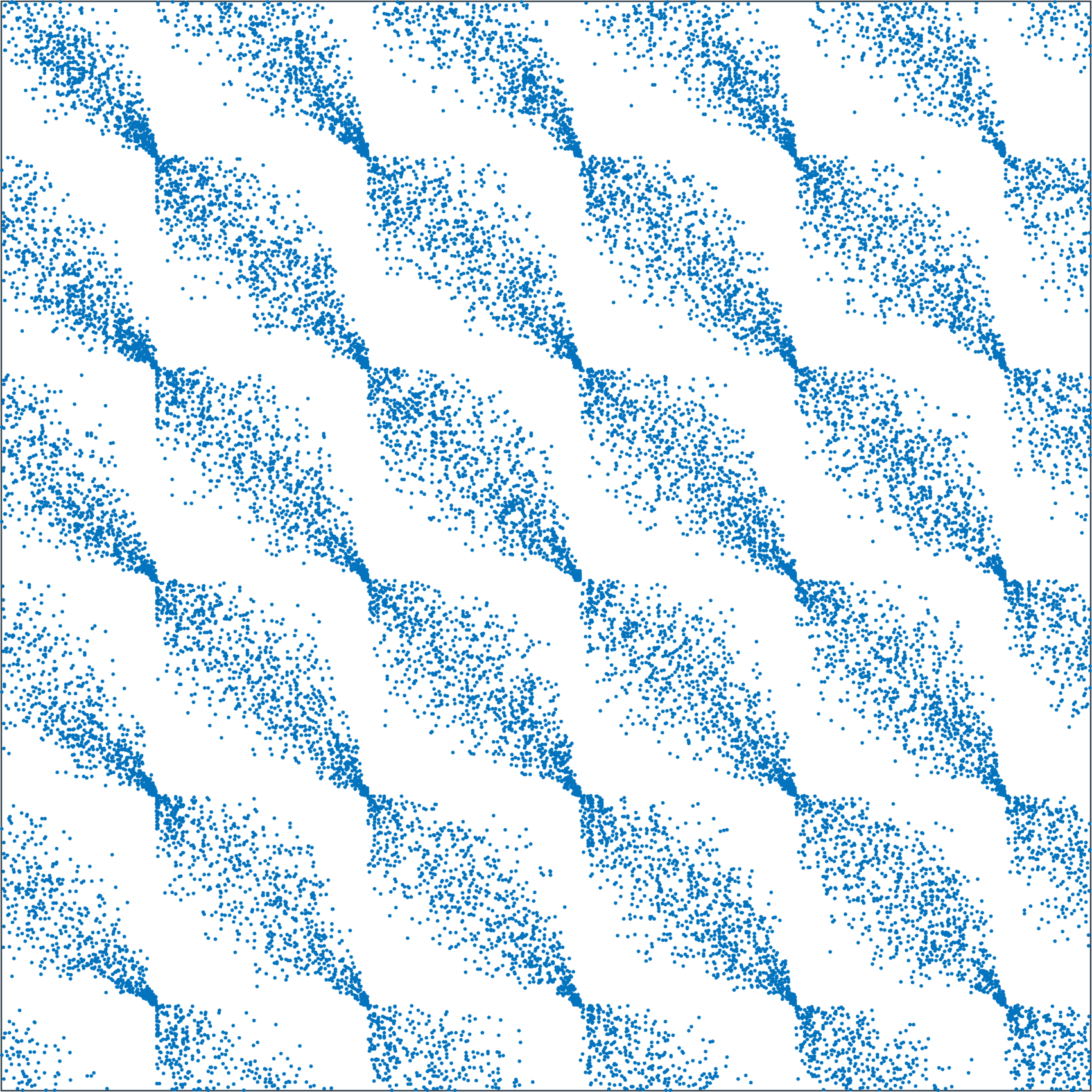%
    }
    \\[1em]
    % 
    % gist
    % 
    \includegraphics[width=\linewidth]{%
      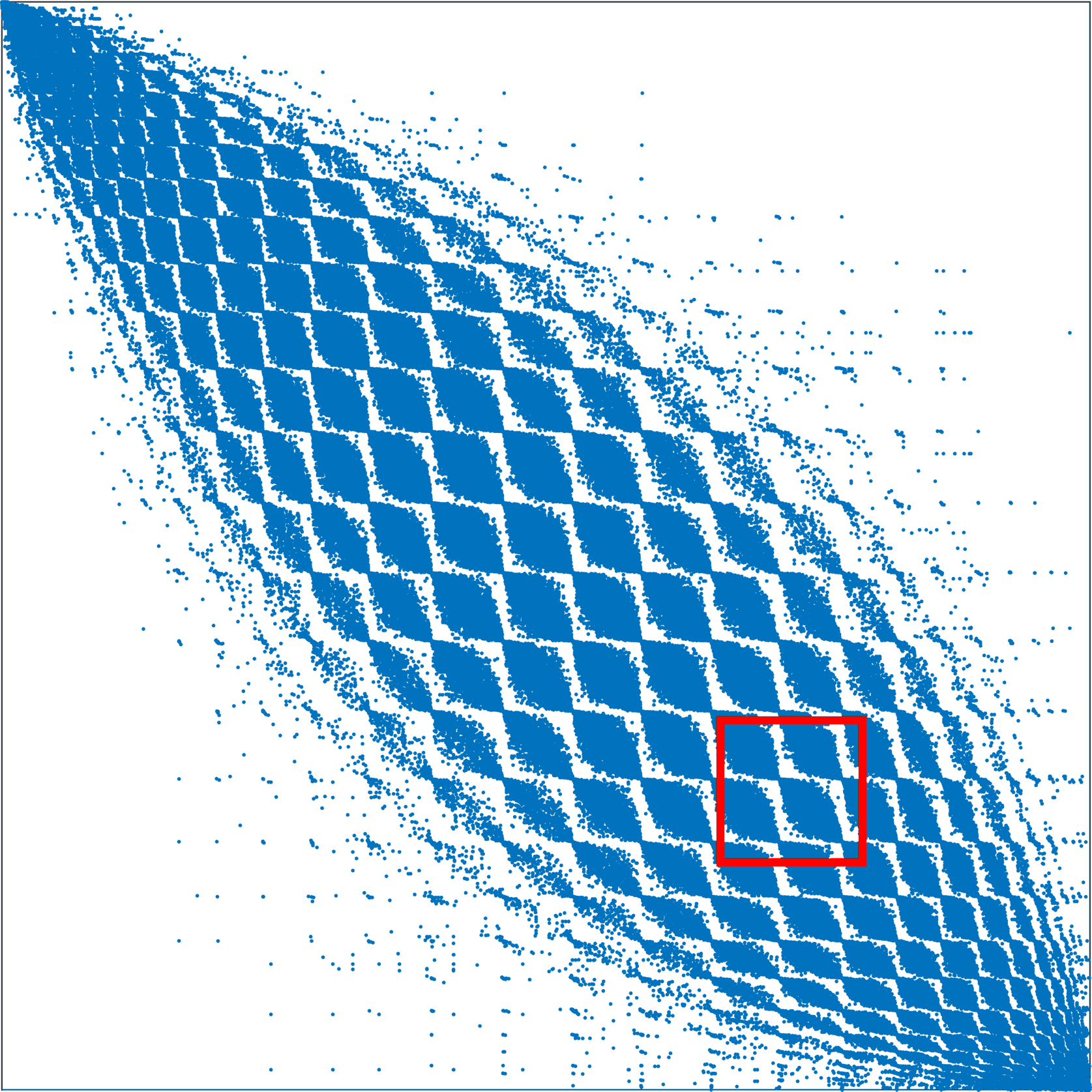%
    }
    \\
    % 
    % gist roi
    % 
    \includegraphics[width=\linewidth]{%
      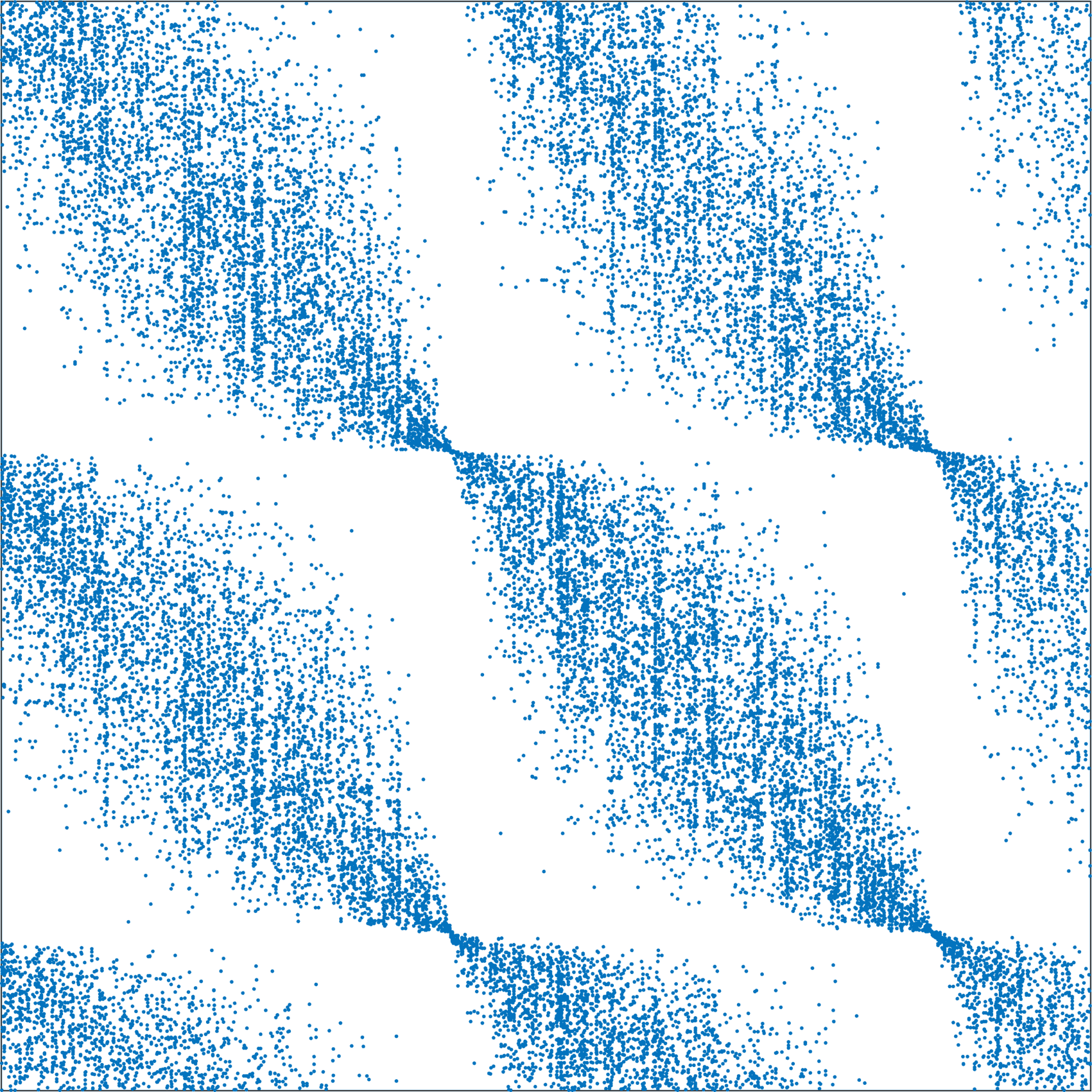%
    }
    \caption{2D lexical}
  \end{subfigure}
  \hspace*{\fill}
  % 
  % 
  % ----- LEXICOGRAPHIC 3D
  % 
  \begin{subfigure}{0.150\linewidth}
    \centering
    % 
    % sift
    % 
    \includegraphics[width=\linewidth]{%
      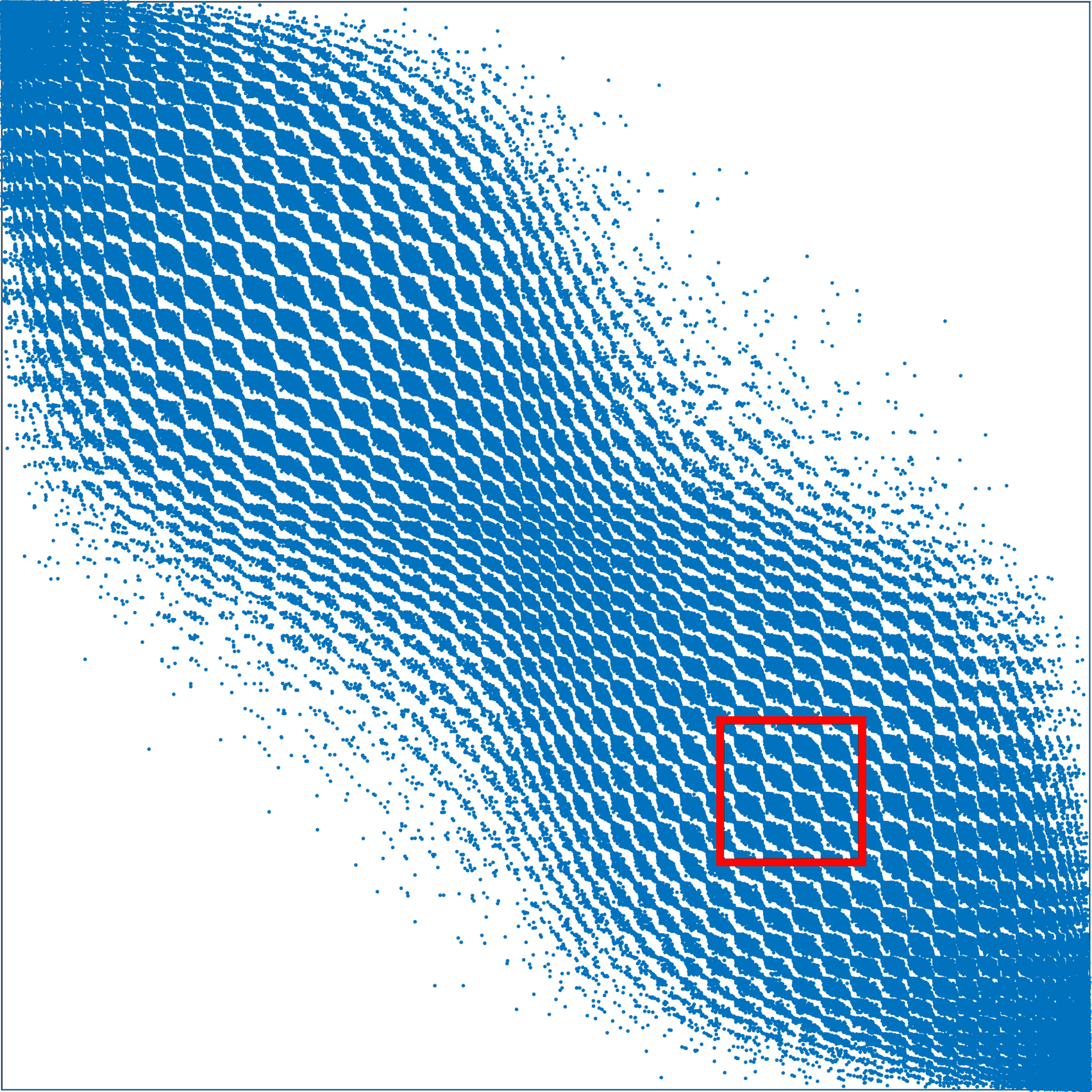%
    }
    \\
    % 
    % sift roi
    % 
    \includegraphics[width=\linewidth]{%
      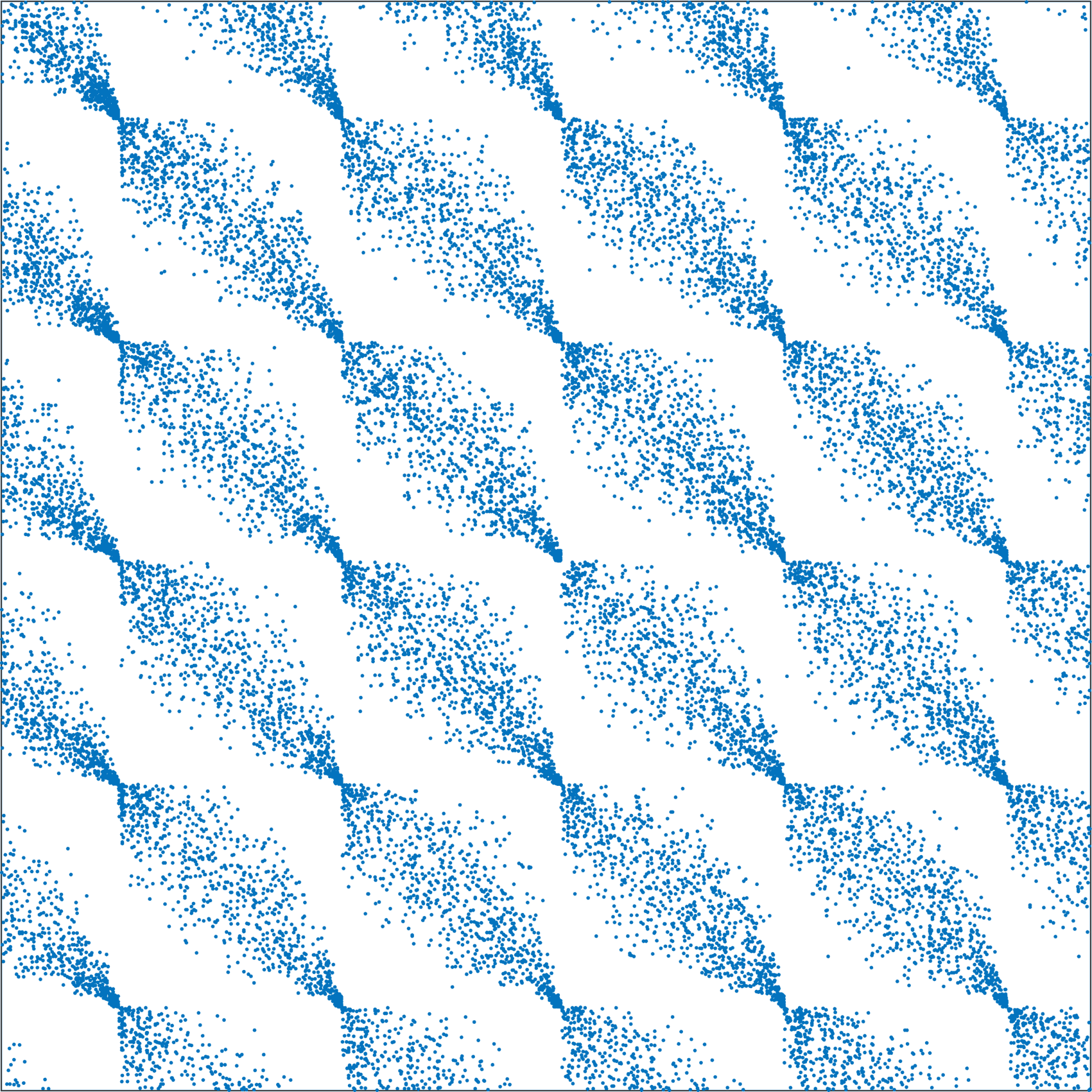%
    }
    \\[1em]
    % 
    % gist
    % 
    \includegraphics[width=\linewidth]{%
      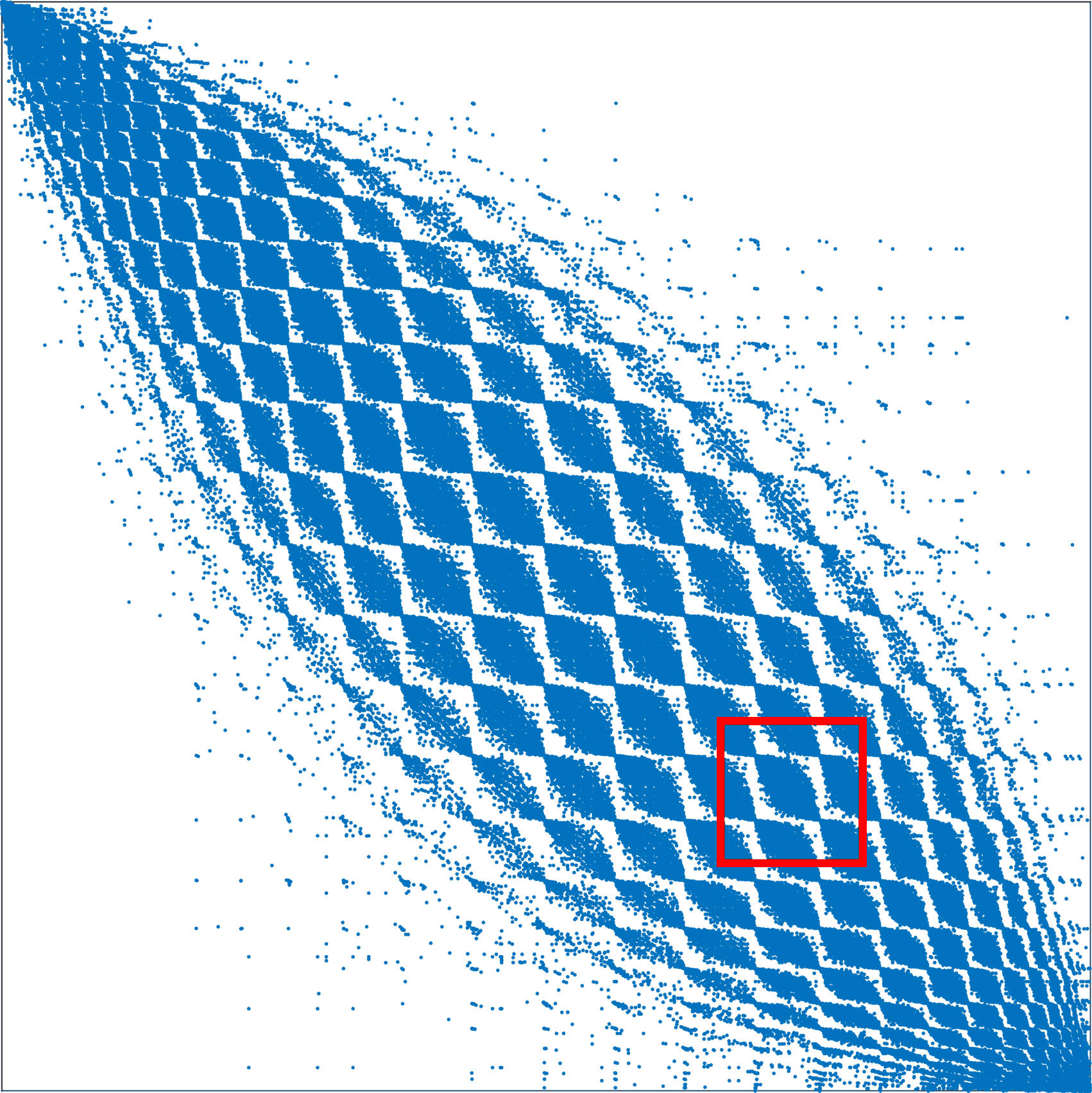%
    }
    \\
    % 
    % gist roi
    % 
    \includegraphics[width=\linewidth]{%
      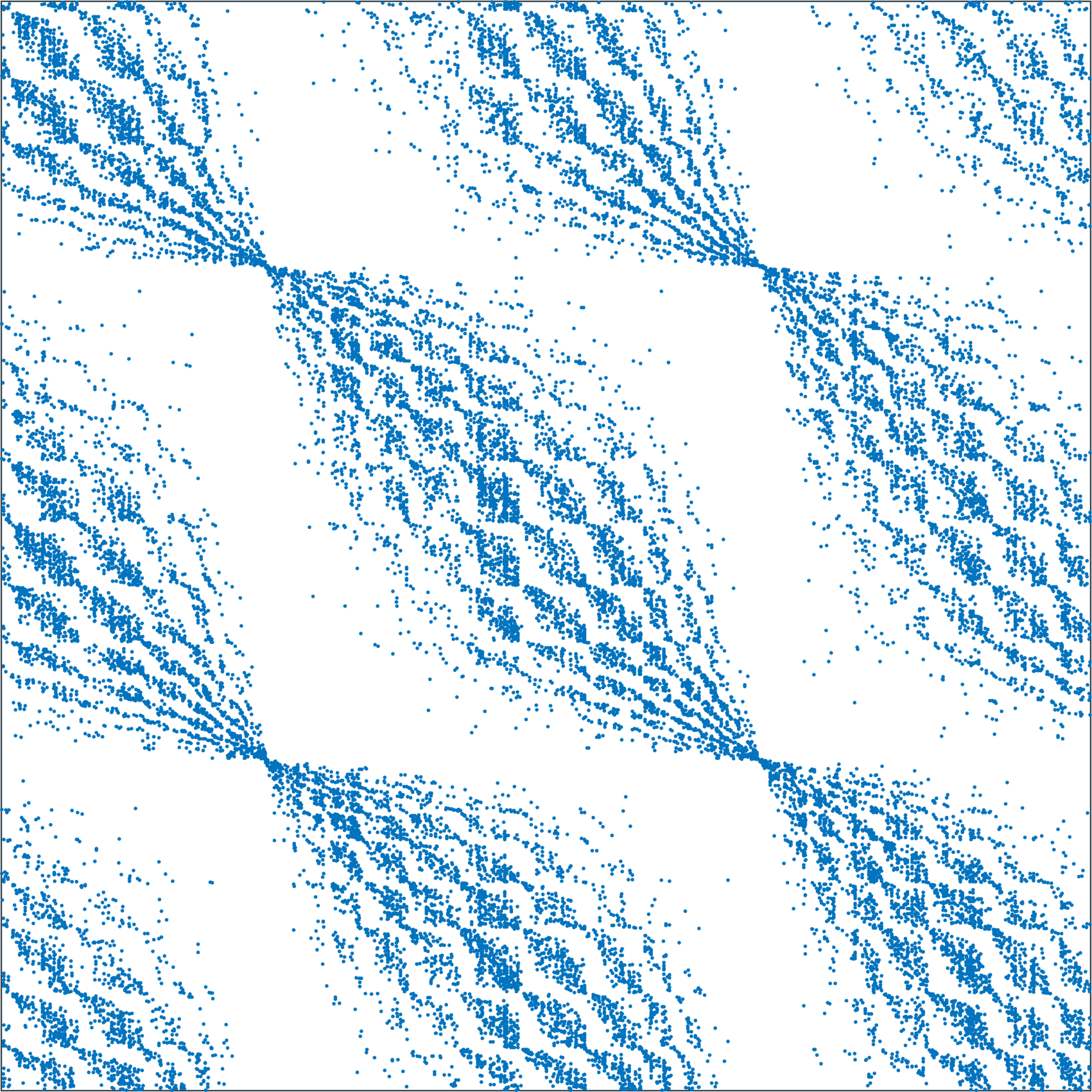%
    }
    \caption{3D lexical}
  \end{subfigure}
  \hspace*{\fill}
  % 
  % 
  % ----- TREE 3D
  % 
  \begin{subfigure}{0.150\linewidth}
    \centering
    % 
    % sift
    % 
    \includegraphics[width=\linewidth]{%
      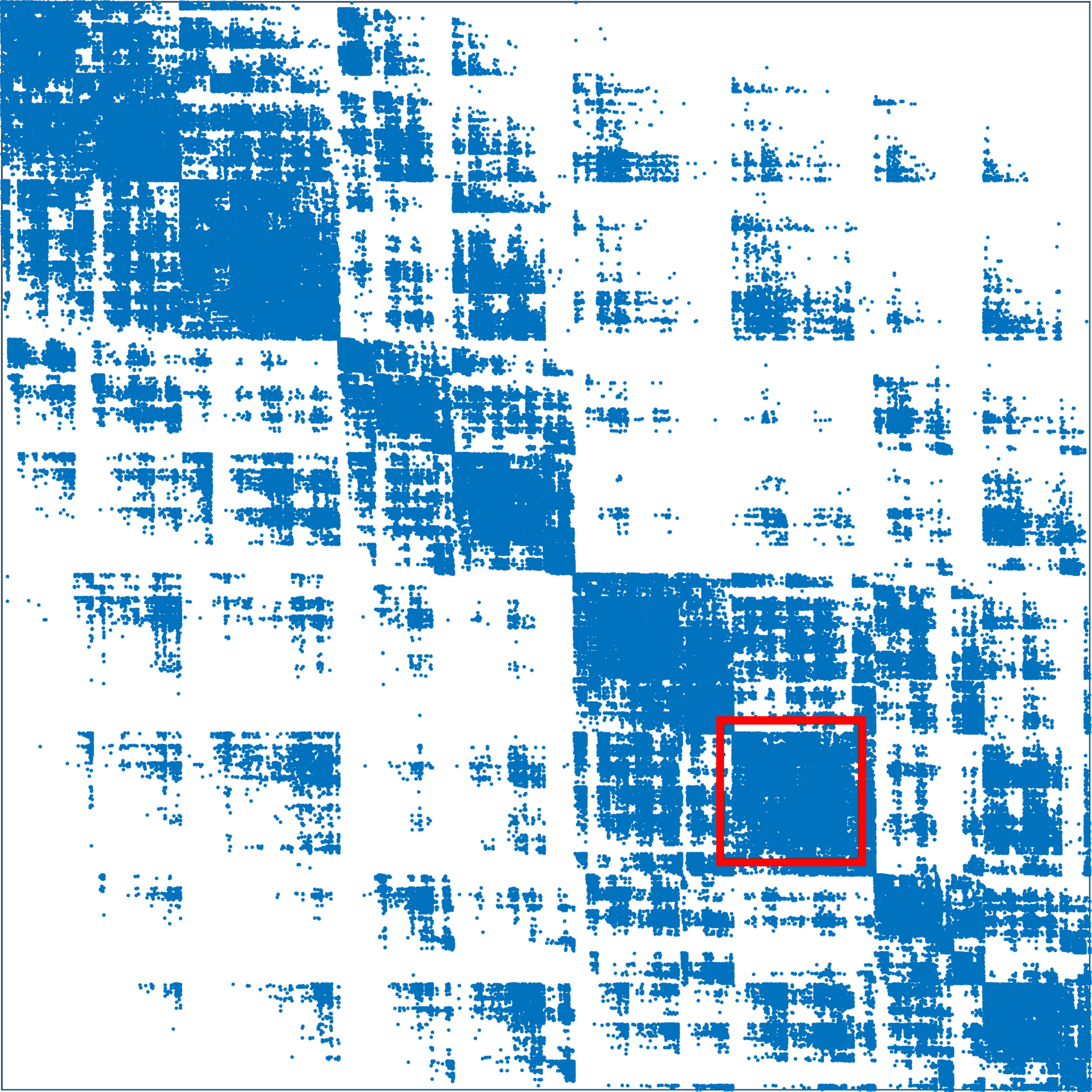%
    }
    \\
    % 
    % sift roi
    % 
    \includegraphics[width=\linewidth]{%
      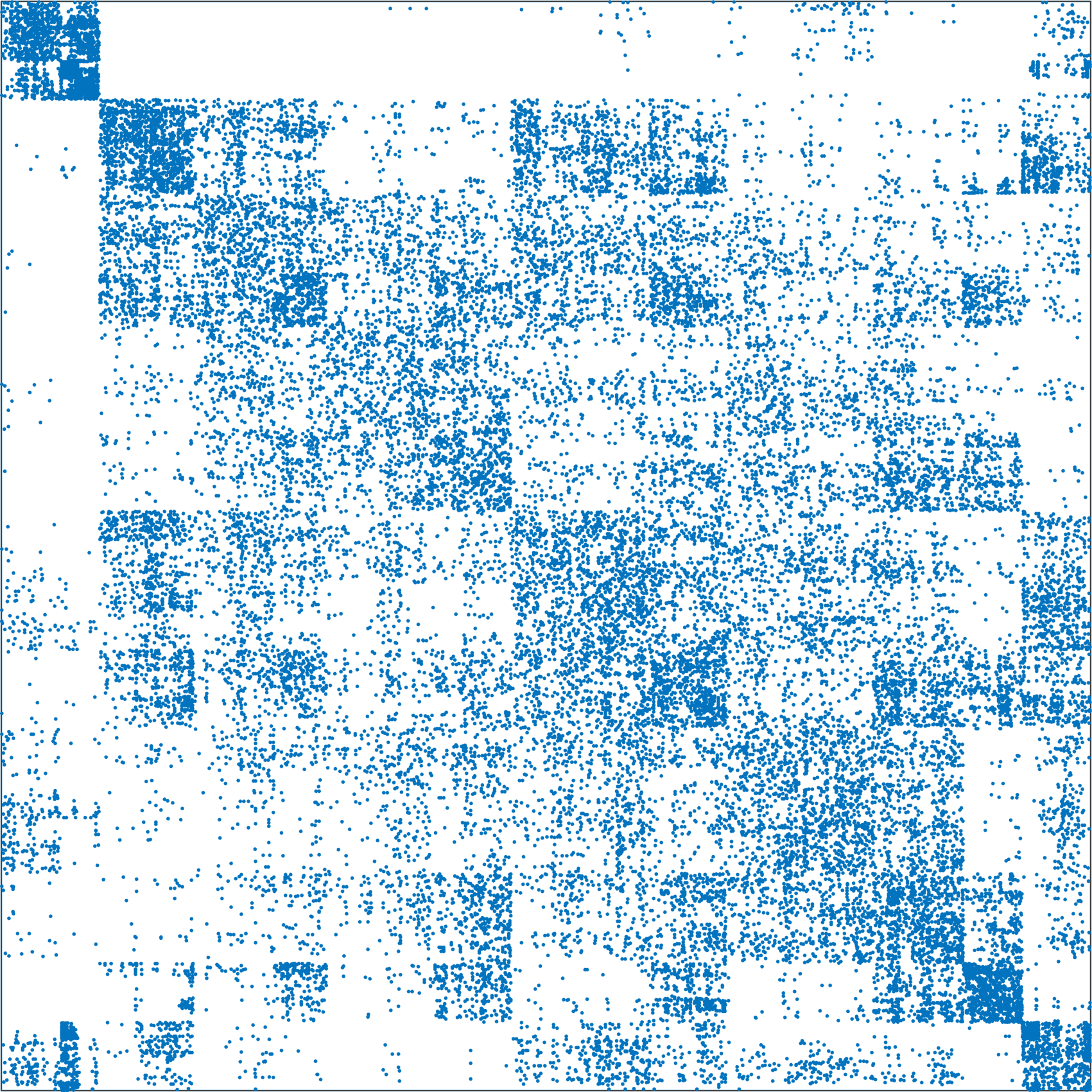%
    }
    \\[1em]
    % 
    % gist
    % 
    \includegraphics[width=\linewidth]{%
      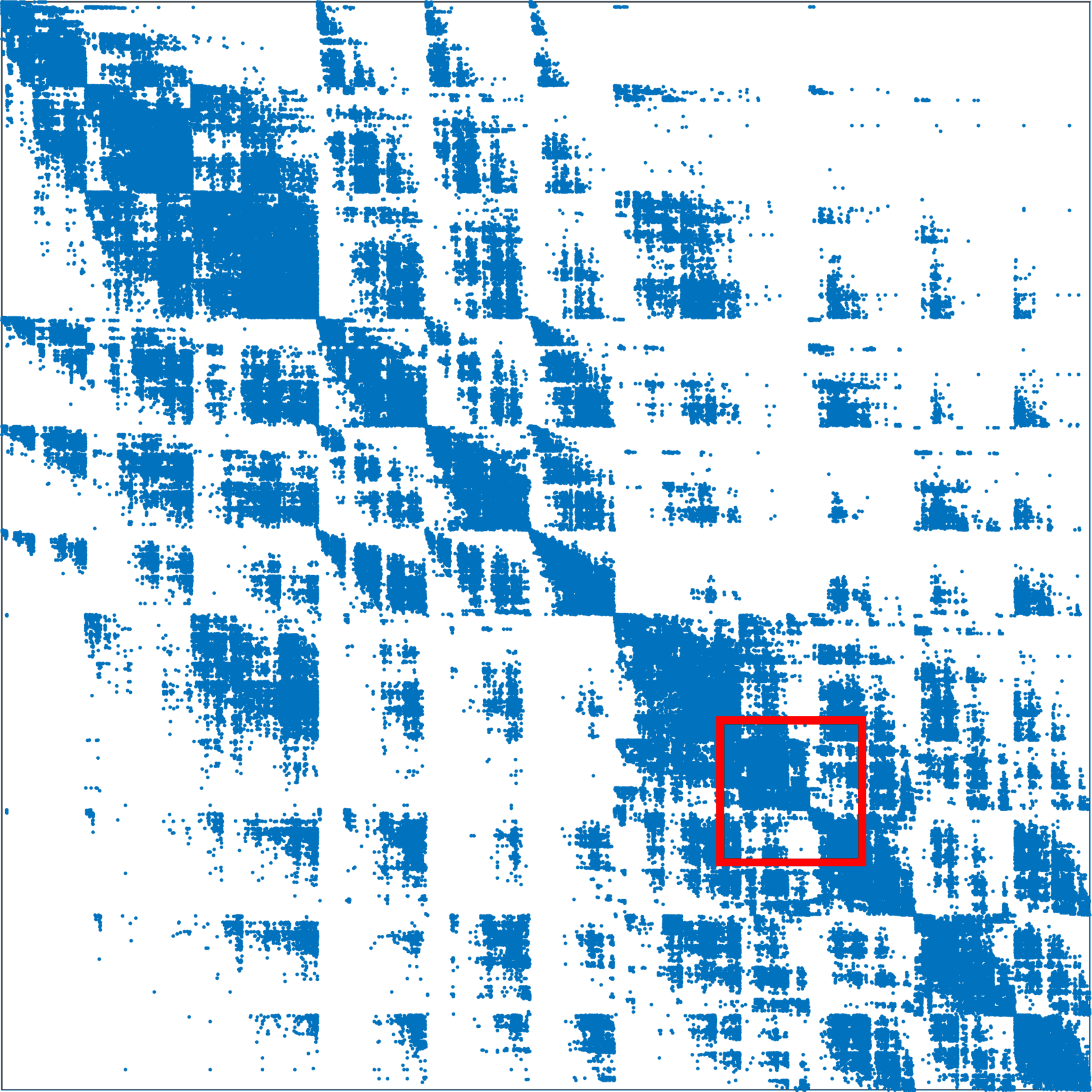%
    }
    \\
    % 
    % gist roi
    % 
    \includegraphics[width=\linewidth]{%
      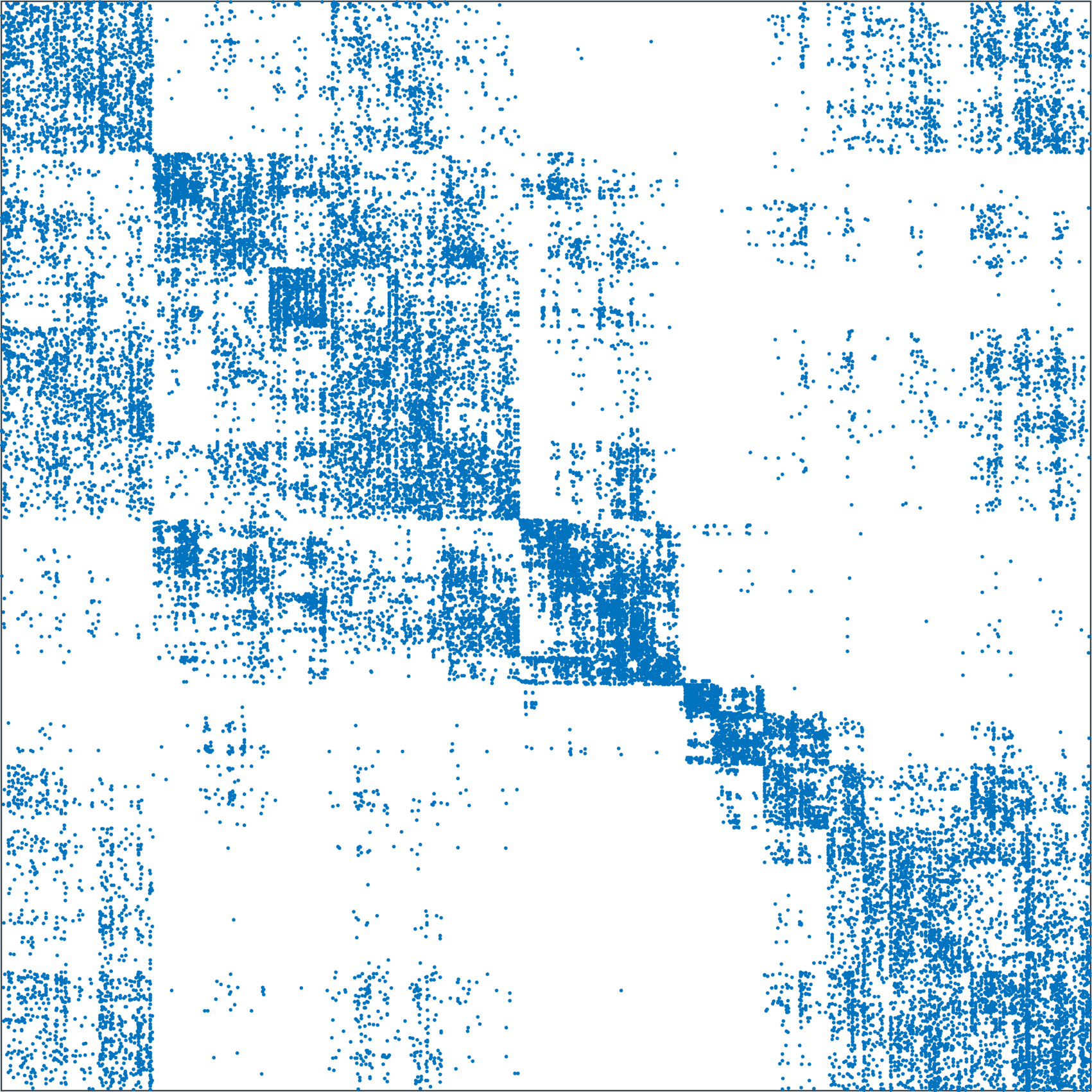%
    }
    \caption{3D dual tree}
  \end{subfigure}
  \caption{Sparse profiles and ROI/sub-matrix details of high-dimensional
    interaction matrices under different orderings.  These matrices capture
    symmetrized interactions between $2^{14}$ randomly selected points from the
    SIFT and GIST datasets, described in Section~\ref{subsec:datasets}. %
    rCM is the reverse Cuthill-McKee ordering~\ccite{george1971}. %
    The 1D ordering refers to sorting the data points by the most dominant PCA
    component coordinates.  Similarly for 2D and 3D lexical ordering, using
    the first 2 or 3 principal components, respectively. %
    The 3D dual tree ordering results from our hierarchical partitioning method,
    described in Section~\ref{sec:matrix-ordering-algorithm}.%
  }
  \label{fig:sparse-profiles}
\end{figure*}

By empirical tests we have carried out, the $\patchDensityKern$-score varies
monotonically with the patch density score over the orderings used in the tests.
See for instance the $\patchDensity$-scores and the $\patchDensityKern$-scores
for the sparsity profiles in Fig.~\ref{fig:example-profiles}.  We show also in
Fig.~\ref{fig:sparse-profiles} the sparse profiles of two real-data interaction
matrices under different orderings, including the ordering by our algorithm,
presented in the following section; the $\patchDensityKern$-score for each one
is listed in Table~\ref{tab:patch-densities}.

% FIGURE: sparsity profiles for experimental interaction matrices
% figure-sparseProfiles.tex

%%% Local Variables:
%%% mode: latex
%%% TeX-master: "MPDO-main"
%%% End:

%  LocalWords:  symmetrized

% TABLE: numerical estimates for different orderings of interaction matrices
% table-patch-densities.tex

\begin{table}
  \centering
  \caption{Kernel-based approximate patch-density estimates for the SIFT and
    GIST dataset interaction matrices, under different orderings, $(\permTgt,
    \permSrc)$, displayed in Fig.~\ref{fig:sparse-profiles}.  Interacting
    neighborhoods comprise $k$ points (varying in order to illustrate the
    difference in the resulting densities).  Patch-density estimates are
    computed with $\scale = k/2$.}
  \label{tab:patch-densities}
  \begin{tabular}{@{} l c d{3.1} d{3.1} d{3.1} d{3.1} d{3.1} d{3.1} @{}}
    \toprule
    \multicolumn{1}{@{}l}{\textbf{Set}}
    & \multicolumn{1}{c}{$k$}
    & \multicolumn{6}{c@{}}{\textbf{Patch density estimate}
      $\patchDensityKern\left( \mat(\permTgt,\permSrc); \scale \right)$}
    \\
    \cmidrule{3-8}
    &
    & \multicolumn{1}{c}{\textbf{rand}}
    & \multicolumn{1}{c}{\textbf{rCM}}
    & \multicolumn{1}{c}{\textbf{1D}}
    & \multicolumn{1}{c}{\textbf{2D lex}}
    & \multicolumn{1}{c}{\textbf{3D lex}}
    & \multicolumn{1}{c@{}}{\textbf{3D DT}}
    \\
    \midrule
    SIFT & 30
    &   2.3   &  14.3 &   6.1 &  12.1 &  12.1 & \bf20.\bf0
    \\
    GIST & 90
    &  71.2   & 243.6 & 286.7 & 352.1 & 361.3 & \bf409.\bf6
    \\
    \bottomrule
  \end{tabular}
\end{table}

%%% Local Variables:
%%% mode: latex
%%% TeX-master: "MPDO-main"
%%% End:

%%% Local Variables:
%%% mode: latex
%%% TeX-master: "MPDO-main"
%%% End:

\subsection{Matrix reordering algorithm}
\label{sec:matrix-ordering-algorithm} 
% matrix-ordering-algorithm.tex

We describe our algorithm for attaining, via matrix reordering, a
sparsity profile that is block sparse with dense blocks. The central
idea to the reordering is to explore and exploit intrinsic cluster
structure in the data.
There are three key components to the algorithm.

\paragraph{Low-dimensional embedding}

Often, data clusters in a high dimensional feature space can be effectively and
efficiently uncovered via a low-dimensional embedding.  We use a nearly
isotropic low-dimensional embedding method.  Specifically, the embedding space
we use is spanned by the most dominant/principal feature axes that are specific
to the data points.
This can be done by an economic-sparse version of the singular value
decomposition~(SVD), namely principal component analysis~(PCA).
The 1D embedding by the most dominant component axis is closely related
to the Laplacian spectral embedding by the Fiedler
eigenvector~\cite{fiedler_algebraic_1973}.  Recursive use of the Fiedler
eigenvectors remains restricted to 1D geometry.  We advocate
multi-dimensional embedding, using more than one principal axes.

Notably, dimension reduction is used in many algorithms for high
dimensional data analysis. In our case study with \tsne, for instance,
the principal feature axes are readily available.  In such cases, the
first step in our algorithm incurs no additional computation cost.  When
the feature dimension $D$ is low already, the embedding step is skipped.

For our specific purpose, a modest embedding dimension serves well in practice.
We will show in particular 2D and 3D embeddings, and their advantage over 1D
embedding.  Formally, the embedding dimension $d<D$ can be determined according
to some tolerance on the distortion in pairwise distances and the geometric
neighbor relations.  The tolerance can be translated into the ratio
$\sum_{i=1}^{d} \sigma_i^2 / \norm{\dataMat}_F^2$, where $\dataMat$ is the
(centered) data feature array, and $\sigma_i$, where $i = 1, \ldots, d$ are the
$d$ largest singular values of $\dataMat$.  The ratio can be easily and
economically obtained, without requiring the computation of all $D$ singular
values.

\paragraph{Hierarchical partitioning}

In the low-dimensional embedding (or feature) space, we partition the
data points hierarchically and adaptively to systematically reveal
inherent cluster structure.  With 3D embedding, for example, we use an
adaptive octree to locate and represent the source clusters at multiple
spatial scales.  Hierarchical clustering of the source data leads to a
multi-level blocking among the columns of the interaction matrix.
Similarly, the target tree leads to a blocking among its rows.  The
result is a hierarchical sparsity profile of the matrix; the profile is
block-sparse with dense blocks between two consecutive layers in the
hierarchy.

We will demonstrate empirically that such hierarchical profiles are
better than those in lexical orderings; and that multi-level clustering
is better than single-level clustering.

\paragraph{Multi-level data structure  and interactions}

In order to exploit the hierarchical matrix structure, we reorder the
charge and potential vectors hierarchically in memory, per their
respective clusters in source and target data.
We employ a multi-level sparse storage format to place and access the
interaction matrix.
Finally, we make the best use of the spatial locality explored and
extracted from the matrix and vector data by arranging the computation
ordering accordingly, i.e., the interaction is calculated at multiple
levels.  Specifically, we access the nonzero matrix elements block by
block; the charge and potential vectors, segment by segment. A
block-segment multiplication corresponds to the interaction between a
source cluster and a target cluster.  In a multi-level setting, each
block-segment multiplication at an intermediate level is further broken
down into subblock-subsegment multiplications at the next finer level.

As with our data partitioning and placement strategy, we will
demonstrate empirically that multi-level computation of interactions
outperforms its single-level counterpart, in both single-core and
multi-core environments.

%%% Local Variables:
%%% mode: latex
%%% TeX-master: "MPDO-main.tex"
%%% End:

%  LocalWords:  octree Fiedler

\section{Case studies}
\label{sec:CaseStudies}
% motivatingApps.tex

Our method is motivated and tested by two important algorithms,
\tsne~\cite{maaten-hinton2008} and mean shift~\cite{Comaniciu2002}, which are
used frequently in machine learning applications.  Iterative near-neighbor
interactions of large data points in multi- or high-dimensional feature spaces
are important building blocks in both algorithms.
In the rest of this section, we briefly describe the relevant part of each of
the algorithms.  Experimental results will be reported in
Section~\ref{sec:Experiments}.

%%% Local Variables:
%%% mode: latex
%%% TeX-master: "MPDO-main"
%%% End:

\subsection{Attractive interaction in \tsne\ gradient}
\label{subsec:tSNEgradient}
% tsne-nn-interaction.tex

The stochastic neighbor embedding~(SNE) algorithm by Hinton and
Roweis~\cite{hinton2003stochastic} embeds a set $\mbs{X}$ of high-dimensional
data into a $d$-dimensional space, where $d$ is much lower than $D$, the
original feature dimensionality.
The embedding is such that neighbor relationships are preserved by a stochastic
approach.  These relationships are cast into conditional probabilities of
neighborhood governed by Gaussian kernels, and are to be preserved in the
low-dimensional embedding space.
In a particular SNE variant by van der Maaten and and Hinton, named \tsne, the
Student t-distribution kernel is used to govern the conditional probabilities in
the lower-dimensional embedding space~\cite{maaten-hinton2008}.  This particular
algorithm has attracted a lot of attention for its application to fascinating
visualization and inspection of high-dimensional data via their 2D or 3D
embedding.  Its applications, however, are severely hindered by computation
latency, even by the accelerated version~\cite{VanDerMaaten_accelerating_2014}.

In \tsne, the embedding point set $\mbs{Y}$ is to be placed in the $d$
dimensional space by iteratively matching the conditional probabilities of
neighborhood in the embedding space to those in the original feature space.  The
matching objective is achieved by minimizing the Kullback-Leibler (KL)
divergence between the two distributions.
There are two terms in the KL gradient calculation at each iteration step, named
as the attractive and repulsive forces.  We focus in this study on the
calculation of the attractive force, which involves near-neighbor interactions.
In a fixed ordering, the sparsity profile of the near-neighbor interaction
matrix remains unchanged over the iteration.  The values of the nonzero
elements, however, vary with iterative estimates of the embedding data
$\mbs{Y}$; the matrix is therefore updated at each iteration step.

%%% Local Variables:
%%% mode: latex
%%% TeX-master: "MPDO-main"
%%% End:

%  LocalWords:  Hadamard

\subsection{Iterative mean shifting}
\label{subsec:meanShift}
% meanShift-iteration.tex

The mean shift algorithm, originally by Fukunaga and Hostetler in
1975~\cite{fukunaga_estimation_1975}, got renewed interest thanks to an
influential paper by Comaniciu and Meer in 2002~\cite{Comaniciu2002}.  It has
been frequently used for non-parametric cluster analysis or mode allocation in
discrete data.  The algorithm locates the density maxima by iterative estimating
and shifting the weighted means of the data points within a neighbor range, via
a kernel function with local support.  A Gaussian kernel is often used.
Calculation of the shift vectors can be viewed as iterative
near-neighbor interactions between the currently estimated means
(targets) and the provided data points (sources), governed by the
kernel.
During the iteration, the sources do not change, the target means
shift.  As a result, the sparsity structure and the numerical values of
the near-neighbor interaction matrix changes with the iteration.  The
data clustering on the target set needs not to be updated as frequently.

%%% Local Variables:
%%% mode: latex
%%% TeX-master: "MPDO-main"
%%% End:

%  LocalWords:  sparsified

\section{Experiments}
\label{sec:Experiments}
% experiments-intro.tex
%
We present experimental results on near-neighbor interaction performance
in this section.
We provide empirical comparisons among matrix orderings by our method
and other existing methods.  
The new method is superior in sequential and parallel execution.

We provide in Table~\ref{tab:cpus} the specifications of two 
workstations used for the experiments. 
One is equipped with an Intel Core i7-6700 CPU (launched Q3'15)
and $32\;$GB of RAM; the other has two Intel Xeon E5540 CPUs (launched Q1'10)
and $48\;$GB of RAM.

\begin{table}
  \centering
  \caption{Specifications of the CPUs used in our experiments. %
    ``Thr'' indicates the number of available virtual cores due to
    hyper-threading.}
  \label{tab:cpus}
  % 
  % table-cpu-specs.tex
%
%
\begin{tabular}{%
  @{}
  l
  d{1.2}
  d{4.0}
  d{2.0}
  d{5.0} @{$\;\,$}
  d{6.0} @{$\;\,$}
  d{4.0} 
  @{}
  }
  \toprule
  \multicolumn{1}{@{}l}{\textbf{CPU}}
  & \multicolumn{1}{c}{\textbf{Clock}}
  & \multicolumn{1}{c}{\textbf{Cores}}
  & \multicolumn{1}{c}{\textbf{Thr}}
  & \multicolumn{1}{c@{$\;\,$}}{\textbf{L1}}
  & \multicolumn{1}{@{}c@{$\;\,$}}{\textbf{L2}}
  & \multicolumn{1}{@{}c@{}}{\textbf{L3}}
  \\
  & \multicolumn{1}{c}{(GHz)}
  & 
  &
  & \multicolumn{1}{c@{$\;\,$}}{(KB)}
  & \multicolumn{1}{@{}c@{$\;\,$}}{(KB)}
  & \multicolumn{1}{@{}c@{}}{(MB)}
  \\
  \midrule
  % 
  % ---------- BARITONE -- BASSOON -- CLARINET -- BONGO
  % 
  $1\times$ Core i7-6700
  & 3.40
  & 4 
  & 8
  & 4 {\times} 32   
  & 4 {\times} 256  
  & 8    
  \\
  %
  % ---------- LINUX21-28
  % 
  $2\times$ Xeon E5540
  & 2.53 
  & 2 {\times} 4
  & 16
  & 8 {\times} 32   
  & 8 {\times} 256  
  & 2 {\times} 8
  \\
  \bottomrule
\end{tabular}%
%
%
%
%%% Local Variables:
%%% mode: latex
%%% TeX-master: "MPDO-main"
%%% End:

  %
\end{table}

%%% Local Variables:
%%% mode: latex
%%% TeX-master: "MPDO-main"
%%% End:

\subsection{Micro-benchmarks}
\label{subsec:blasBenchmarks}
% blasBenchmarks.tex
%
% Sept. 2017

We make a few benchmarks of near-neighbor interactions on the testbed machines
with synthetic data to establish machine-specific references for performance
assessment.  We use the Intel high-performance \spmvMkl\ implementation of the
BLAS \spmv\ for the benchmarks.

The making of the benchmarks is as follows.  For a fixed matrix size with a
fixed number of nonzeros, we use the banded matrix, which corresponds to 1D
interaction, to establish the best performance obtainable on the particular
machines.  To establish the base case, we use a matrix with the nonzero elements
randomly scattered.  In both cases, the matrix elements are in compressed
storage format and referenced via indirect addresses.  The benchmarks with the
synthetic data are to be used in Section~\ref{subsec:tsneBenchmarks} as a
reference for assessing the performance of near-neighbor interactions with
real-world datasets described in the next section.

%%% Local Variables:
%%% mode: latex
%%% TeX-master: "MPDO-main"
%%% End:

%  LocalWords:  BLAS

\subsection{Datasets}
\label{subsec:datasets}
% datasets.tex
%

The data sets used in the experiments are drawn from the following two
big datasets, which are publicly available: 
\begin{description}
\item[SIFT] 128-dimensional SIFT feature vectors~\cite{Lowe2004}
  extracted from images of the INRIA Holidays
  dataset~\cite{jegou_embedding_2008}.
\item[GIST] 960-dimensional GIST feature
  vectors~\cite{oliva2001modeling} extracted from images of the tiny
  image set~\cite{torralba_million_2008}.
\end{description}

%%% Local Variables:
%%% mode: latex
%%% TeX-master: "MPDO-main"
%%% End:

% FIGURE: t-SNE speed-up
% figure-tsne-speedup.tex

\begin{figure*}
  % 
  % ---------- SIFT on Core i7-6700 [BARITONE]
  % 
  % ----- Y-label
  % 
  \begin{subfigure}{0.01\linewidth}
    \centering
    \vspace*{-5em}
    \rotatebox[origin=c]{90}{\scriptsize Speed-up over sequential scattered}
    \\[8em]
    \rotatebox[origin=c]{90}{\scriptsize Speed-up over sequential scattered}
  \end{subfigure}
  % 
  % ----- plot
  % 
  \begin{subfigure}{0.48\linewidth}
    \centering
    \hspace*{\fill}
    \includegraphics[width=0.98\linewidth]{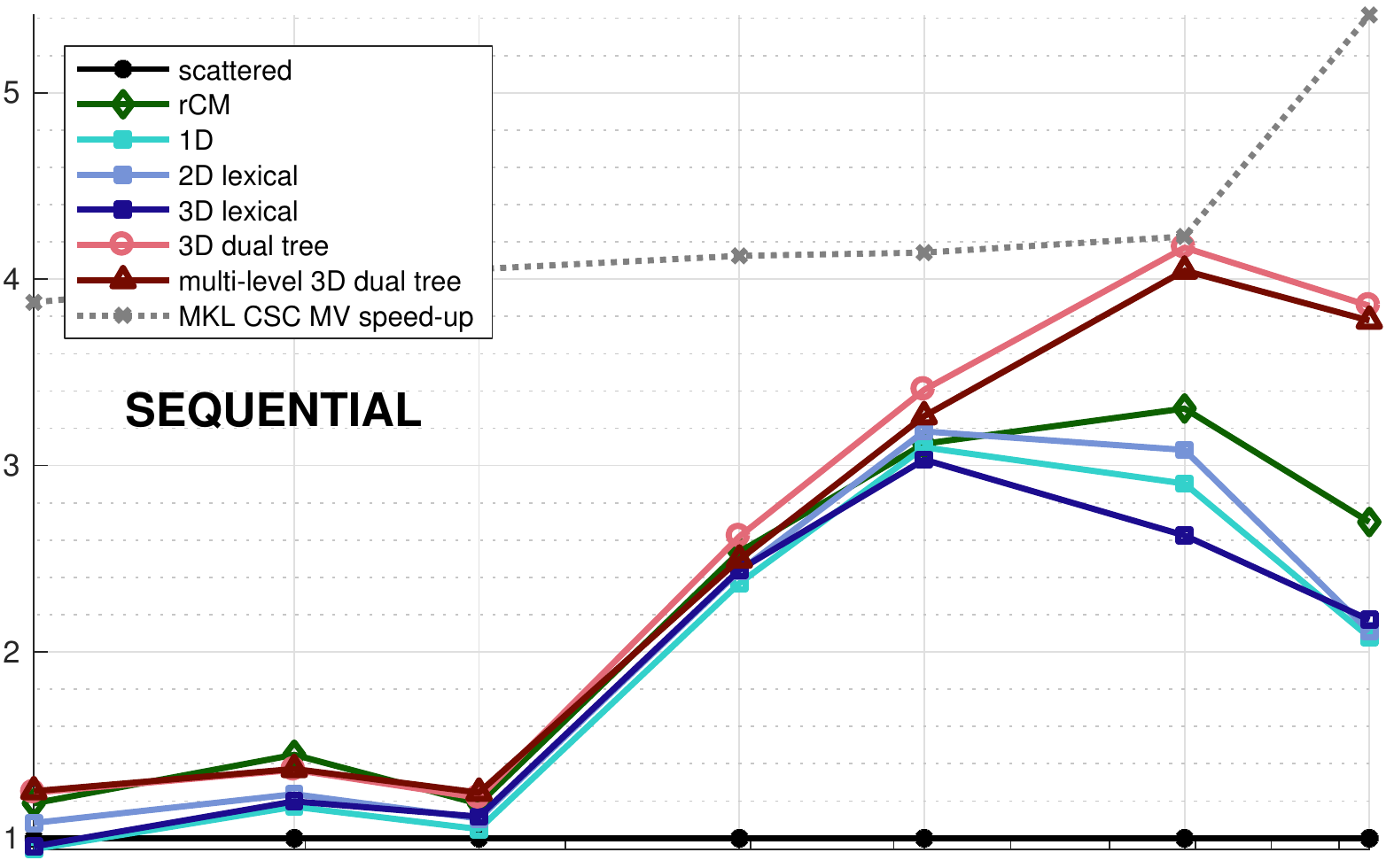}
    \\[0.5em]
    \hspace*{\fill}
    \includegraphics[width=0.99\linewidth]{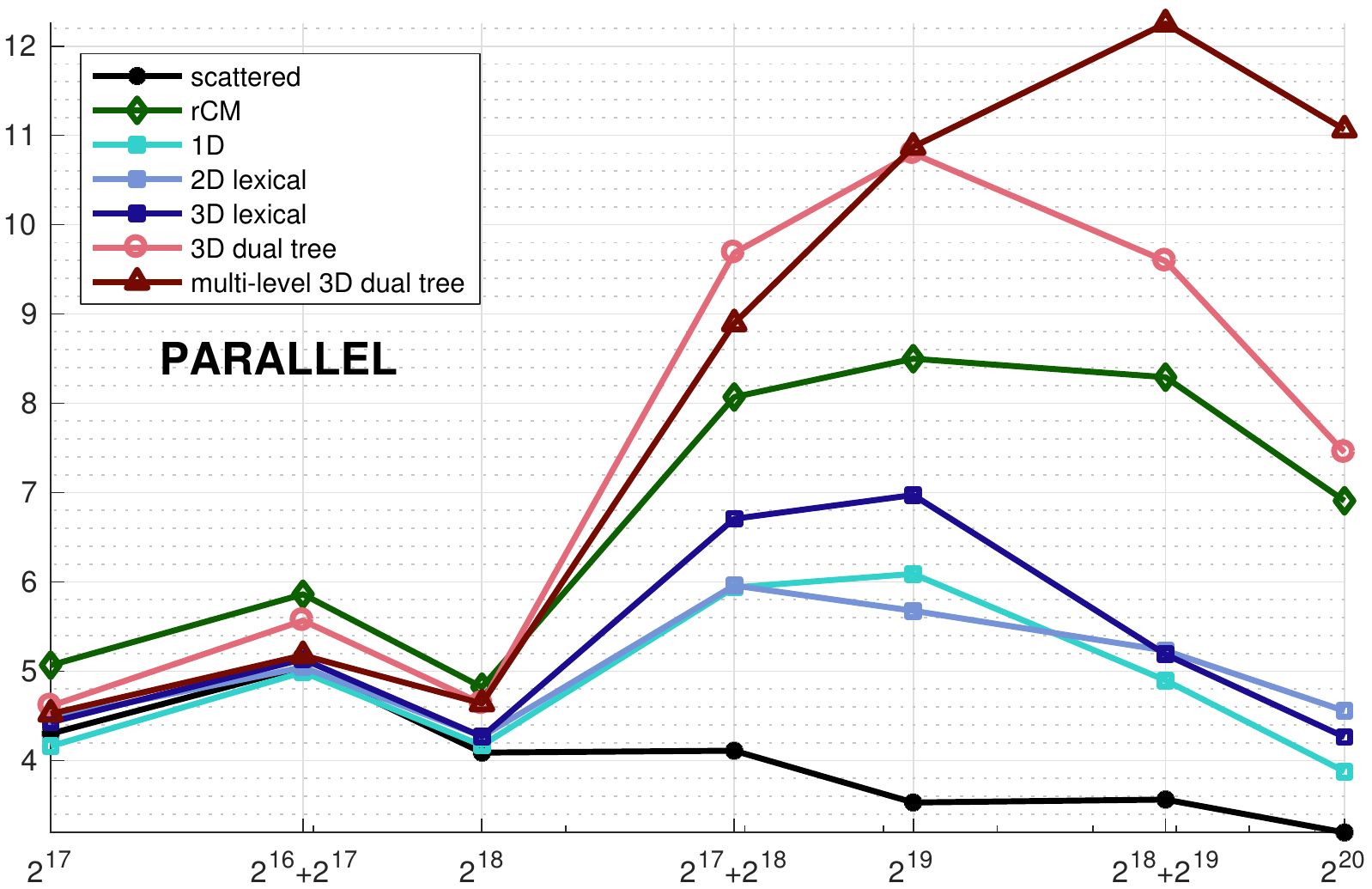}
    \\
    {\scriptsize Number of data points}
    \caption{SIFT dataset; Core i7-6700 workstation}
    \label{fig:tsne-benchmarks-a}
  \end{subfigure}
  \hspace*{\fill}
  % 
  % ---------- GIST on Xeon E5540 [LINUX{21-28}]
  % 
  \begin{subfigure}{0.48\linewidth}
    \centering
    \hspace*{\fill}
    \includegraphics[width=0.98\linewidth]{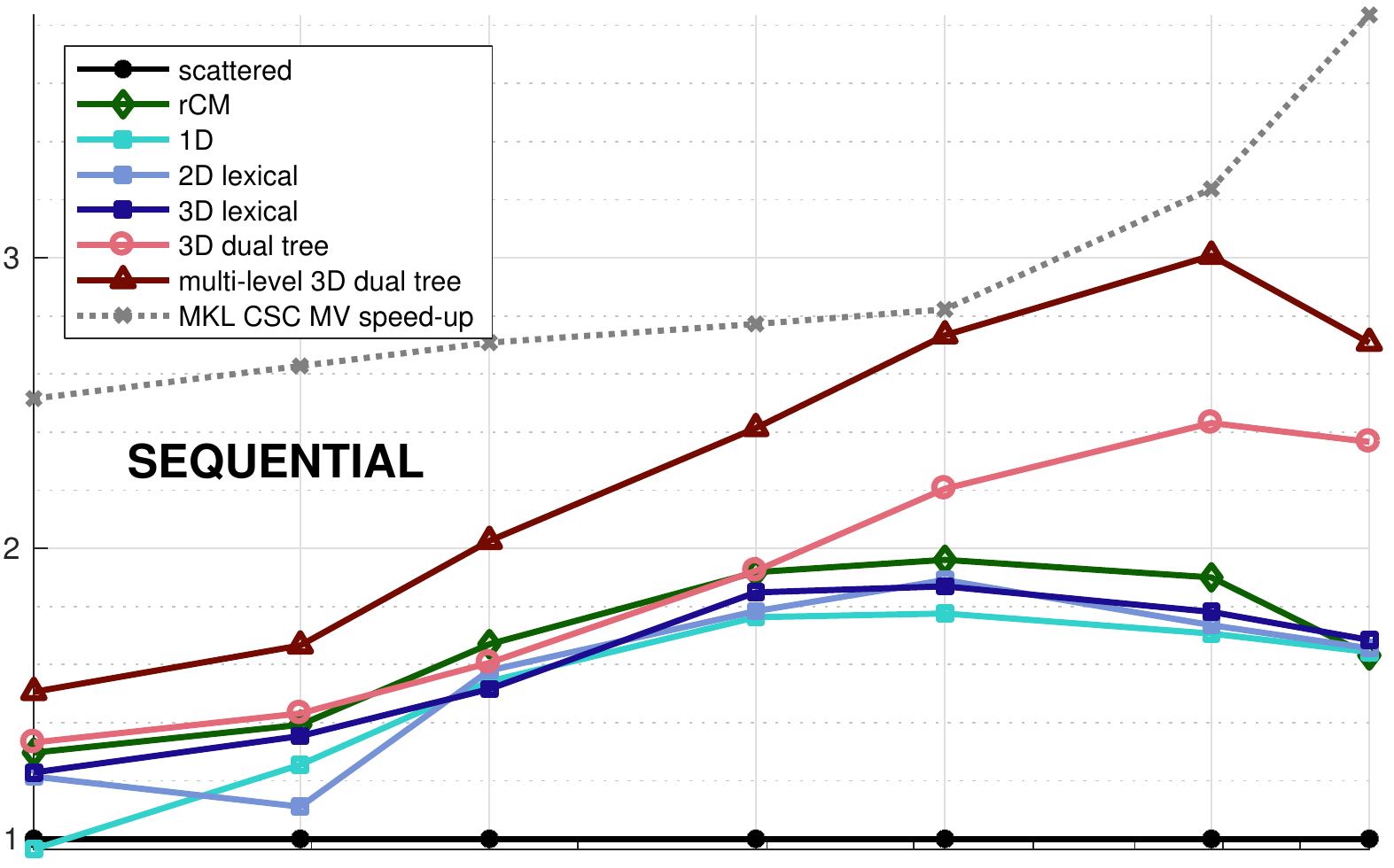}
    \\[0.5em]
    \hspace*{\fill}
    \includegraphics[width=0.99\linewidth]{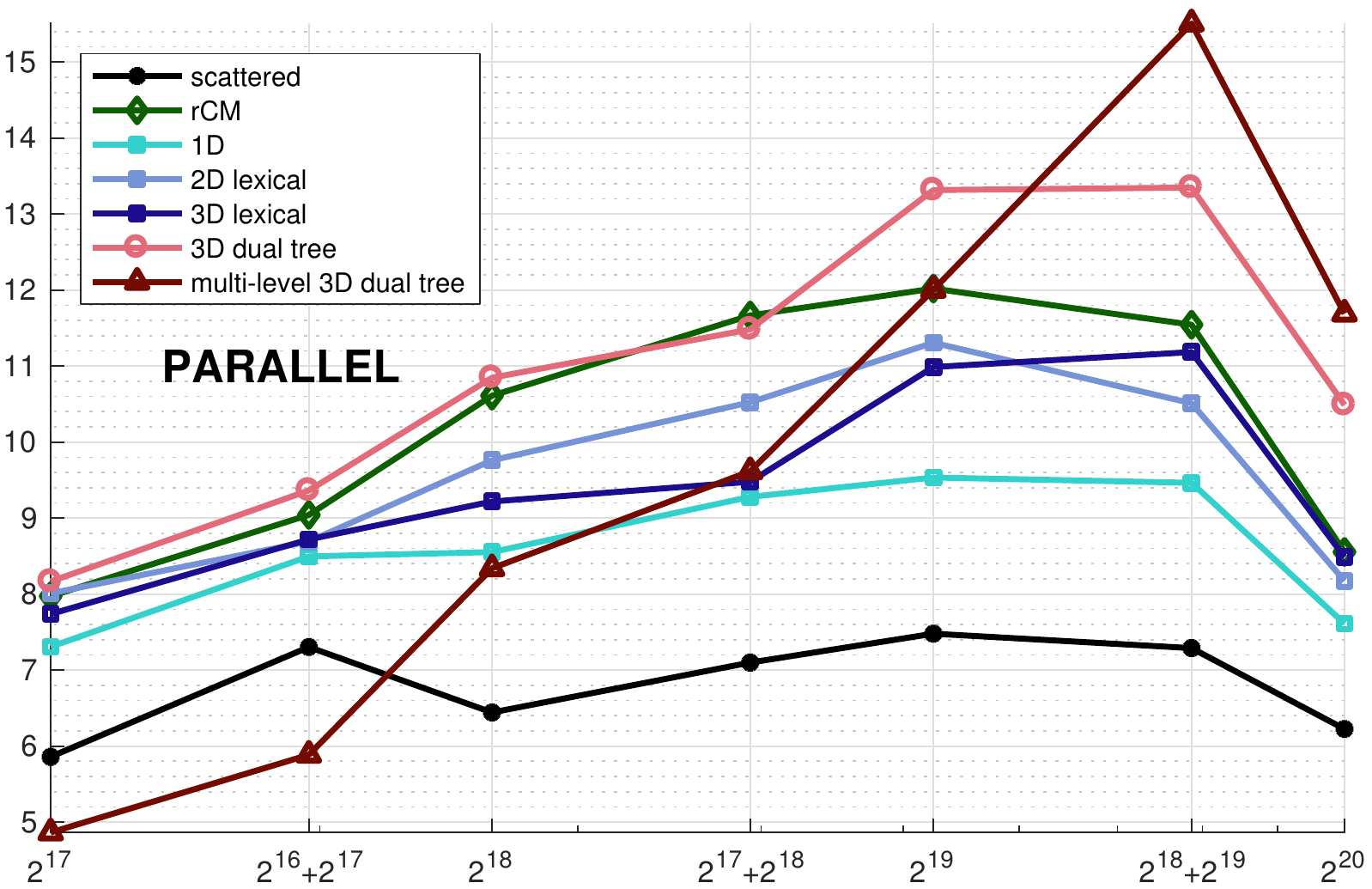}
    \\
    {\scriptsize Number of data points}
    \caption{GIST dataset; Xeon E5540 workstation}
    \label{fig:tsne-benchmarks-b}
  \end{subfigure}
  \caption{%
    Comparison in performance by different ordering schemes used for attractive
    force calculation in \tsne.  The top plots show the comparison in sequential
    execution; the bottom, in parallel.  The plots to the left are associated
    with the SIFT data on the Core i7-6700 workstation; to the right, with the
    GIST data on the Xeon E5540 workstation.  The solid black line in each case
    corresponds to the case with randomly scattered ordering.  The sequential
    execution time is used as the reference time, for comparisons in sequential
    execution as well as parallel execution.  More details are in the text. %
  }
  \label{fig:tsne-benchmarks}
\end{figure*}

%%% Local Variables:
%%% mode: latex
%%% TeX-master: "MPDO-main"
%%% End:

\subsection{Performance}
\label{subsec:tsneBenchmarks}
% tsneBenchmarks.tex

We provide in Fig.~\ref{fig:tsne-benchmarks} the comparison in
performance among several ordering schemes used for the attractive force
calculation in \tsne. The comparisons are made on both data sets, in
sequential execution as well as in parallel execution.  The reference
time for the comparisons is the sequential execution time with the
scattered ordering.

The following particular orderings are used in the comparisons:
\begin{inparaenum}
\item ``scattered,'' by random permutation of the interacting points placement;
\item ``rCM,'' the reverse Cuthill-McKee ordering;
\item ``1D,'' where the data points (rows and columns) are sorted by the most
  dominant PCA component coordinates;
\item ``2D lexical'' and
\item ``3D lexical,'' by lexicographic sorting of the first 2 or 3 principal
  components, respectively; and
\item ``3D dual tree,'' by our matrix reordering algorithm,
  described in Section~\ref{sec:matrix-ordering-algorithm}.
\end{inparaenum}
The sparse matrix profiles associated with each of orderings are shown in
Fig.~\ref{fig:sparse-profiles}.

The dotted gray lines in the top plots show the time ratio of the
execution time of \spmvMkl\ as per the micro-benchmarks discussed in
Section~\ref{subsec:blasBenchmarks} (speed-up of banded over scattered
matrix operations).  The number of nonzero elements per row is constant
and matches the sparsity of the SIFT and GIST near-neighbor interaction
matrices on the corresponding workstation experiments.
These time ratios are used as a reference for the maximum expected improvement
to be gained by matrix reordering.

The speed-up in parallel execution by our new method reaches
$12 {\times}$ on the Core i7 and more than $15 {\times}$ on the Xeon
E5540.  Compared to the popular rCM ordering, the multi-level 3D dual
tree ordering is about 40\% faster for larger problem sizes, when the
data size exceeds the size of the L3 cache.
On both workstations and datasets, the sequential execution improvement
with our reordering approaches the \spmvMkl\ time ratio for certain
sizes.

We underline the following observations.
By multi-dimensional embedding, our method explores more potential in data
clustering; this leads to better matrix sparsity profiles, improved data
locality, reduced memory latency, and hence fast execution time.
The performance with hierarchical ordering outperforms that with the lexical
orderings in the same embedding space.  This shows the importance of integrating
multi-dimensional embedding with hierarchical clustering.  The latter is
followed by multi-level data placement, and multi-level interactions.
The comparisons are consistent with what we expected from the respective 
sparity profiles in Fig.~\ref{fig:sparse-profiles}.

%%% Local Variables:
%%% mode: latex
%%% TeX-master: "MPDO-main"
%%% End:

\section{Related work \& discussion}
\label{sec:related-work-discussion}
% relatedWork.tex

Improving data locality in sparse matrix computation by reordering the
matrix at the algorithm level has a long and rich history.  It becomes
more important with modern computers and computation applications.
The CM ordering by Cuthiil and McKee was reported in
1969~\cite{cuthill1969}; the reverse CM~(rCM) ordering appeared two
years later~\cite{george1971}.  Davis et al.\ provided in 2016 a
survey of fill-reducing orderings in direct methods for sparse linear
systems~\cite{davis2016survey}.

Our method assumes that the sparse matrix is defined over coordinated data and
represents near-neighbor interactions.  In particular, the feature vectors serve
as coordinate vectors.  Exploiting coordinate attributes in the data was
discussed in~\cite{Mellor-Crummey2001} and the references therein.  The
dimension of the coordinate space was low.  In modern data and image analysis,
the feature dimension is typically much higher.  High dimensionality exposes the
limitation of previous sparsity profiles.  For example, the size of the
bandwidth envelope is essentially a 1D measure, relying on a 1D embedding of
multi-dimensional data.  In particular, 1D Laplacian spectral embedding by the
Fiedler vector~\cite{fiedler_algebraic_1973} is frequently used in previous work
to reduce the size of bandwidth envelope~\cite{barnard1995}.  We advocate
multi-dimensional embedding, instead.

Multi-dimensional embedding serves two objectives simultaneously: near-isometric
dimension reduction, and multi-level clustering in the lower-dimensional space.
The final goal is to attain a desirable matrix sparsity profile.  Our method
does not invoke the formation of a Laplacian graph when it is not readily
available.  The recursive bisection method by the Fiedler vectors of partitioned
subgraphs may be viewed as multi-level clustering.  It is, however, limited to
1D embedding geometry.  Our multi-level partition is in a multi-dimensional
space, without entailing computation of recursive Fiedler vectors.

Our assumption and method are directly applicable to a broad class of data and
operations for data analysis.  This class includes mesh data in various
scientific simulations; it includes graph or network data with attributes on the
graph nodes and/or edges.  Investigation of graph-related sparse compression and
computation is reported in Chapter~8 of Shun's dissertation in
2017~\cite{shun2015}.

Our multi-level compressed sparse storage format has a connection to the
Compressed Sparse Block~(CSB) scheme by Bulu{\c{c}} et
al.~\cite{bulucc2009parallel}.  CSB makes, stores, and accesses uniform blocks
of a preset size, assuming no knowledge of data coordinates and cluster
structure.  By such blocking, the distance between successive accesses to the
entries in the same matrix block is bounded constant from above, away from the
growing data size.  Our scheme reduces to CSB when the hierarchy is flat, i.e.,
with only a single level of blocks, except that the blocks at the bottom level
are more or less uniform in the number of nonzeros, but not necessarily uniform
in block area.  At a higher level in our compression hierarchy, the blocks are
pointers for indirect references to hierarchically placed data.

The temporal ordering, or partial ordering, in computational execution must be
compatible with the spatial ordering, relationship, and placement of data.
Although well noted in previous work, this spatio-temporal compatibility
warrants special attention and effort in algorithm development and
implementation.

In summary:
\begin{inparaenum}
\item Our method for ordering a sparse matrix by multi-scale clustering of
  high-dimensional data is based on a novel concept and model of desirable
  sparsity profiles.  Our block-sparse with dense blocks model and the related
  patch density measure capture and characterize the essential properties in an
  interaction matrix that are conducive to better space and time locality.  The
  model and measure unify several previous sparsity profile models in the sense
  that our model favors the same profile favored by another one when the data
  meet the condition(s) by the latter.
\item Our method is empirically shown to be superior to previous, popular
  methods in sequential computation on a single core, and even better in
  parallel computation on multiple cores.
\end{inparaenum}

%%% Local Variables:
%%% mode: latex
%%% TeX-master: "MPDO-main"
%%% End:

%%%%%%%%%%%%%%%%%%%%%%%%%%%%%%%%%%%%%%%%%%%%%%%%%%
%%% END OF MAIN DOCUMENT

% ---------- ACKNOWLEDGMENT

\begin{acks}
  We gratefully acknowledge an equipment grant by Intel.
\end{acks}

% ---------- BIBLIOGRAPHY

\bibliographystyle{ACM-Reference-Format}

\bibliography{ref}

\end{document}